\documentclass[letterpaper, 10 pt, conference]{ieeeconf}  % Comment this line out if you need a4paper

\IEEEoverridecommandlockouts                              % This command is only needed if 
\usepackage{amsfonts}
\usepackage{subcaption}
\usepackage{graphicx}
\usepackage{booktabs}

\usepackage{comment}

\usepackage[ruled,vlined]{algorithm2e}

\title{\LARGE \bf
Approximating Constraint Manifolds Using Generative Models for Sampling-Based Constrained Motion Planning
}

\author{Cihan Acar and Keng Peng Tee% <-this % stops a space
\thanks{The authors are with the Institute for Infocomm Research (I$^2$R), A*STAR, Singapore 138632.} 
\thanks{Email: \small \{acar\_cihan, kptee\}@i2r.a-star.edu.sg}%
}

\begin{document}

\maketitle
% \pagenumbering{arabic}
% \thispagestyle{plain}
% \pagestyle{plain}

%%%%%%%%%%%%%%%%%%%%%%%%%%%%%%%%%%%%%%%%%%%%%%%%%%%%%%%%%%%%%%%%%%%%%%%%%%%%%%%%
\begin{abstract}

Sampling-based motion planning under task constraints is challenging because the null-measure constraint manifold in the configuration space makes rejection sampling extremely inefficient, if not impossible. This paper presents a learning-based sampling strategy for constrained motion planning problems. We investigate the use of two well-known deep generative models, the Conditional Variational Autoencoder (CVAE) and the Conditional Generative Adversarial Net (CGAN), to generate constraint-satisfying sample configurations.  Instead of precomputed graphs, we use generative models conditioned on constraint parameters for approximating the constraint manifold. This approach allows for the efficient drawing of constraint-satisfying samples online without any need for modification of available sampling-based motion planning algorithms. We evaluate the efficiency of these two generative models in terms of their sampling accuracy and coverage of sampling distribution. Simulations and experiments are also conducted for different constraint tasks on two robotic platforms.

\end{abstract}

%%%%%%%%%%%%%%%%%%%%%%%%%%%%%%%%%%%%%%%%%%%%%%%%%%%%%%%%%%%%%%%%%%%%%%%%%%%%%%%%
\section{INTRODUCTION}

Robot motion planning is concerned with finding a collision-free continuous path, from a start configuration to a goal configuration, which satisfies a set of constraints such as joint limits, self-collision, and kinematic constraints~\cite{LaValle}. For high-dimensional complex robotic systems, sampling-based motion planning algorithms can be effectively implemented in high-dimensional configuration spaces (C-space). Sampling-based algorithms work by sampling random valid configurations and forming a collision-free graph or tree of valid motions using these valid configurations. Probabilistic Road Maps (PRM) and Rapidly-exploring Random Trees (RRTs)~\cite{Steven1998} are two of the most popular sampling-based motion planning algorithms often shown to yield good performance in practice~\cite{Simic2014}.

Collision constraints alone are relatively easy to deal with using rejection sampling since the volume of collision-free configuration space is usually not negligible (exceptions in e.g. narrow corridor problem). In contrast, task constraints are generally difficult to ensure when using sampling-based motion planning. For tasks with pose constraint or closed-chain kinematic constraint, there is a very low, if not null, probability of randomly sampling a constraint-satisfying configuration, due to the null-measure manifold  induced by these task constraints in the configuration space \cite{LaValle11}. Therefore, rejection sampling is usually not feasible for highly constrained motion planning problems.
%, which makes them challenging to solve with the rejection based method. 

Constrained motion planning can be done by tailoring the method to only specific task constraints, such as orientation constraint, closed-chain kinematic constraint, and view constraint, among others. However, this approach is not general for handling new or different task constraints, and it might be difficult to combine different methods for handling multiple simultaneous task constraints present in certain applications. 

\begin{comment}
 \begin{figure}[t]
    \centering
    \begin{subfigure}{0.23\textwidth}
        \includegraphics[width=\textwidth]{images/Olivia_box.png} 
        % \caption{}
        % \label{fig:hook1}
    \end{subfigure}
    \begin{subfigure}{0.23\textwidth}
        \includegraphics[width=\textwidth]{images/Pholus_reach.png} 
        % \caption{}
        % \label{fig:hook2}
    \end{subfigure}
    \caption{\small Challenging tasks requiring constraints to be satisfied during entire motion plan. Dual-arm object manipulation (left) involves closed-chain kinematic constraints, and whole-body reaching (right) balance constraints. }
    \label{fig:hook}
\end{figure}
\end{comment}

In this paper, we focus on using generative models to sample valid configurations on constraint manifolds, motivated by the success of generative models in producing some of the most realistic content in different domains including image, video, and text, among others. 
We employ two of the most popular and powerful generative models, VAE and GANs, to approximate the constraint manifolds for pose-constrained manipulation tasks and a balance-constrained whole-body motion task. Task-independent environment constraints, such as obstacles, are not entangled in our learned models. Thus, the trained models can be reused in different/new environments with the same task specifications (e.g. different obstacle fields). Furthermore, the learning-based framework can easily approximate any new task constraint manifold by prior training on relevant data, and can be paired with any motion planner that receives the constraint-satisfying configuration samples from the generative model.          

The contributions of this paper are:
\begin{itemize}
    \item[i)]  A method for sampling-based constrained motion planning using generative models, learned from prior/demonstrated task data, to approximate the constraint manifold and generate new constraint-satisfying samples online/offline for constrained motion planning.  
    \item[ii)] Performance comparison analysis between two of the most popular generative models, CVAE and CGAN, and insight on the accuracy of generating constraint-satisfying configurations and coverage of sampling distribution. 
    \item[iii)] Validation on different task constraints, including end-effector pose constraint, closed kinematic chain constraint, and balance constraint to show the generality and flexibility of the approach to different applications and environments. 
\end{itemize}

\section{Related Work}
\label{sec:Background}
To solve the constrained motion planning problem, various techniques have been proposed, including ones based on projection, tangent space, atlas construction, and inverse kinematics (IK) (see \cite{Kingston2018} for a review). In projection-based methods, randomly sampled  configurations are projected onto the constraint manifold by using iterative techniques. Since the early works on randomized gradient descent approach \cite{Lavalle99}, variants of the projection strategy have been developed, including random linear projection \cite{Sucan09} and the popular Jacobian pseudoinverse projection \cite{Stilman07,Berenson2011}. To reduce the number of projection operations, planning on the tangent bundle of the constraint manifold was proposed \cite{Suh11}. An atlas for the constraint manifold was built online and incrementally on which planning was carried out directly \cite{Porta12}. 

IK-based methods try to satisfy the task space constraints via IK for a subset of joints so that constraint-free planning can be performed on the remainder joints.
To deal with closed kinematic chains, methods such as kinematics-based roadmap \cite{Li00} and Random Loop Generator \cite{Cortes04} decompose closed chains into open subsets, sampling from a subset and using IK to ensure that closure constraints are satisfied for the rest of the joints. Recently, IK-based motion planning for a dual-arm robot with orientation constraints was proposed in \cite{Wang19}. Based on some assumptions on the kinematic structure of the arms, an analytic IK solution was obtained for the wrist joints such that the orientation constraints were strictly satisfied. However, each of these methods was tailored to a specific task constraint (orientation or closed-chain kinematic constraints) considered in the papers and would not generally apply to other types of task constraints (e.g. balance constraints) or a combination of constraints.

Instead of coming up with a new planner for constrained motion planning, Sucan~\cite{Sucan2012} used offline precomputed approximation graphs to approximate the constraint manifold. In this technique, a set of constraint-satisfying configurations is generated offline, stored as edges in the graph, and then drawn during the planning process to solve the constrained motion planning problem. The advantages of this technique are that constraint-satisfying configurations can be sampled very fast, and there is no need to make any changes to the planner. They showed the reduction of planning time for orientation constraint and dual-arm constraint on different environments using the PR2 robot with respect to the projection-based methods. Similar offline sampling was also applied for whole-body motion planning on a humanoid robot to sample configurations that satisfy balancing constraints \cite{Burget2013}. The biggest issue with the approximation of constraint manifolds with the precomputed dataset is that it is not probabilistically complete because sampling is limited by the fixed dataset, and the manifold coverage is determined by these finite numbers of samples. It also has the drawback of inflexibility against constraints which can be parameterized by more than a few parameters. 

Recently, learning-based approaches are proposed for motion planning ~\cite{Ichter2018}-\nocite{Ichter2019, Chen2020, Zhang2018, Qureshi2019}\cite{Kumar2019}. Learning sampling distributions from demonstrations using CVAE is recently proposed to improve the performance and exploration strategies of sampling-based motion planning algorithms~\cite{Ichter2018}. To achieve that, CVAE is trained conditioned on the initial state, goal region, and obstacles in the environment. A similar method also used CVAE trained on a prior database  to predict roadmaps for sampling-based motion planning problems~\cite{Kumar2019}. The issue with these approaches is that since learning is based on previous demonstrations or data to bias the samples toward promising regions, performance is highly affected for new environments which might have different obstacle configurations. In this work, we utilize deep generative models to estimate the sampling distribution of constrained robot configurations. Instead of using training data that depend on obstacles in the environment, we train our models on robot configuration data parameterized by task constraints. To the best of our know knowledge, this is the first study that investigates the feasibility of using generative models for sampling constraint-satisfying configurations in constrained motion planning problems.

\section{Approximating Constraint Manifolds Using Generative Models}
\label{sec:ACM}
 Generative models are powerful tools for learning representations of high dimensional data~\cite{Hu2017}, and are able to learn complex constraint manifolds. In this paper, we utilize them to approximate the constraint manifolds for constrained sampling-based motion planning. We propose to sample constraint-satisfying configurations using two of the most popular generative models -- VAE and GAN -- instead of sampling from a fixed database which contains a limited number of samples. Using generative models, on the other hand, makes  it possible to generate infinitely many samples and results in greater coverage of the constraint manifold. To handle parameterized constraints, we use conditional versions of VAE and GAN. Our approach focuses on the sampling part and is decoupled from the planning algorithms. Therefore, it is possible to use it directly with available sampling-based motion planning algorithms. 

\subsection{Conditional Variational Autoencoders (CVAE)}
\label{sec:CVAE}	
Variational Autoencoders (VAEs) are powerful generative models for learning latent (hidden) representations. In our case, we utilize them to approximate the constraint manifold and generate new constraint-satisfying configurations for constrained motion planning problems. An autoencoder mainly consists of two components: i) an encoder which compresses the input data into a latent vector referred to as a ‘bottleneck’, and ii) a decoder to reconstruct the original data using this latent vector. The latent vector is regularized to generate new data using the decoder. Training of the VAE is based on maximizing the following variational lower bound:
\begin{eqnarray}\label{eq:elb}
  \mathcal{L}(\theta,\phi;x) =&&\hspace{-6mm}-\beta D_{KL}(q_\phi(z|x)||p_\theta(z)) + \nonumber\\
                              &&\hspace{-6mm}\mathbb{E}_{z \sim q_\theta(z|x)}[log_{p_\theta}(x|z)]
\end{eqnarray}
where the first term (KL divergence) is considered as a regularizer and the second term reconstruction loss. To control the effect of regularization, the weighting factor $\beta$ in the range of 0 to 1 is also introduced~\cite{Higgins2017}. In this equation, $p_\theta(z)$ represents the decoder and $q_\phi(z|x)$ the encoder. After training, the encoder can be discarded, and the decoder used to generate new data. 
 
In the conditional version of VAE,  both latent variable and input data are conditioned to some random variable $y$, which in our case, represents the constraint parameters. For instance, condition variables for end-effector orientation constraint can represent the roll, pitch, and yaw values of the constraint. Similarly, condition variables for position constraint can represent the \textbf{x}, \textbf{y}, and \textbf{z} values. By changing the values of the condition variable, we can generate samples that satisfy different constraints. To generate new constraint-satisfying configurations, the latent vector is sampled from the unit Gaussian and then used as an input to the trained decoder. The output of the decoder is a novel robot configuration that can be used by the motion planner. 

\subsection{Conditional GANs (CGANs) }
\label{sec:CGANs}
Generative adversarial networks (GANs) use two neural networks, competing against each other to generate new data~\cite{Goodfellow2014}. The first network is called a \emph{generator} which tries to generate new data similar to the original, and tries to convince the second network, called a \emph{discriminator}, that the generated data is a real one. The discriminator, on the other hand, tries to discriminate whether the data from the generator is real or not. Both generator and discriminator are trained simultaneously. The original GAN is a two-player min-max game with value function $V(D,G)$, defined as:  
\begin{eqnarray}
    \label{eq:minimax_value}
    \min_G \max_D V(D,G) = && \hspace{-6mm}\mathbb{E}_{x \sim p_{data}}[\log D(x)] + \nonumber\\
                       && \hspace{-6mm}\mathbb{E}_{z \sim p_z(z)}[\log(1 - D(G(z)))]
\end{eqnarray}
where $D(x)$ represents the discriminator and  $G(z)$ the generator. During training, the discriminator tries to maximize this function and, at the same time, the generator tries to minimize it. 

Similar to CVAEs, conditional GANs (CGANs) have been proposed to generate conditional samples~\cite{Mirza2014}, based on the following objective function for CGANs: 
\begin{eqnarray}
\label{eq:minimax_value_condition}
\min_G \max_D V(D,G) =&& \hspace{-6mm}\mathbb{E}_{x \sim p_{data}}[\log D(x|y)] + \nonumber\\
                      && \hspace{-6mm}\mathbb{E}_{z \sim p_z(z)}[\log(1 - D(G(z|y)))]
\end{eqnarray}
Here, both the generator and discriminator are conditioned to some random variable $y$. Similar to CVAEs, the condition variable $y$ for CGANs represents the parameters of the constraints and makes it possible to generate configurations satisfying parameterized constraints with the same model.

\subsection{Generating Samples from the Constraint Manifold}
To generate new constraint-satisfying configurations from CVAE, a latent vector $z$ is sampled from the normal distribution and used as an input to the decoder of the CVAE  with condition variable $y$. For CGAN, noise (latent) vector $z$ is sampled from a uniform distribution and then used as an input to the generator of the CGAN with condition variable $y$. In our case,
% $y$ presents the constraint parameters, and 
the decoder/generator is trying to generate robot configurations that satisfy the constraints. We present the process of generating new constraint-satisfying configurations in Algorithm \ref{alg_prqa}.

\begin{algorithm}
	{\small
 	    \SetKwInOut{KwOff}{Offline}
  	    \SetKwInOut{KwOn}{Online}
		\SetKwInOut{KwIn}{Input}
		\SetKwInOut{KwOut}{Output}
		\caption{Generate Samples from Model}
		\label{alg_prqa}
		\KwOff{}
		Collect constraint-satisfying configurations $q$ using IK. \\
		Train generative model $M_{gen}$ (CVAE or CGAN) conditioned on constraint parameters $y$.  \\
		\KwOn{}
	     Initialize new planning problem $\{q_{start} ,q_{goal}, y\}$.\\
         Generate $N$ new samples from $M_{gen}$, conditioned on $y$, and push them inside a buffer Queue.\\
		\While{Planning}
		{
			\While{Queue is Not Empty }
	        {
		       Pop a sample $q_{new}$ from Queue.\\
		       \If{ $q_{new}$ satisfies constraints}
	                {return $q_{new}$}
	        } 
            Generate $N$ new samples from $M_{gen}$, conditioned on $y$, and push them inside buffer Queue.
		}
	}
	\vspace*{-1mm}
\end{algorithm}

\begin{figure}[ht]
    \centering
    \begin{subfigure}{0.36\textwidth}
        \includegraphics[width=\textwidth]{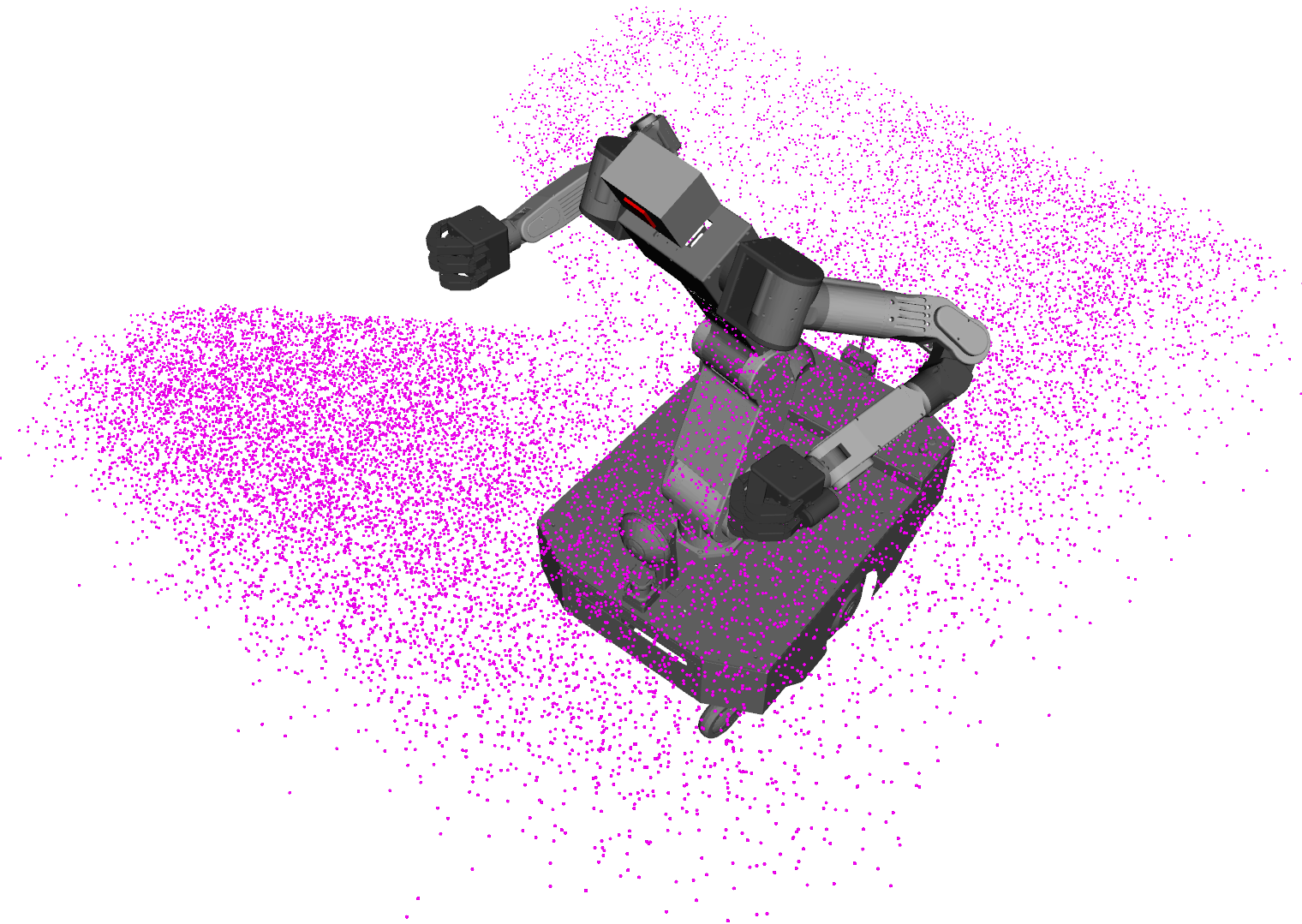} 
        \caption{CVAE}
        \label{fig:manifold_app_CVAE}
    \end{subfigure}
    \begin{subfigure}{0.34\textwidth}
        \includegraphics[width=\textwidth]{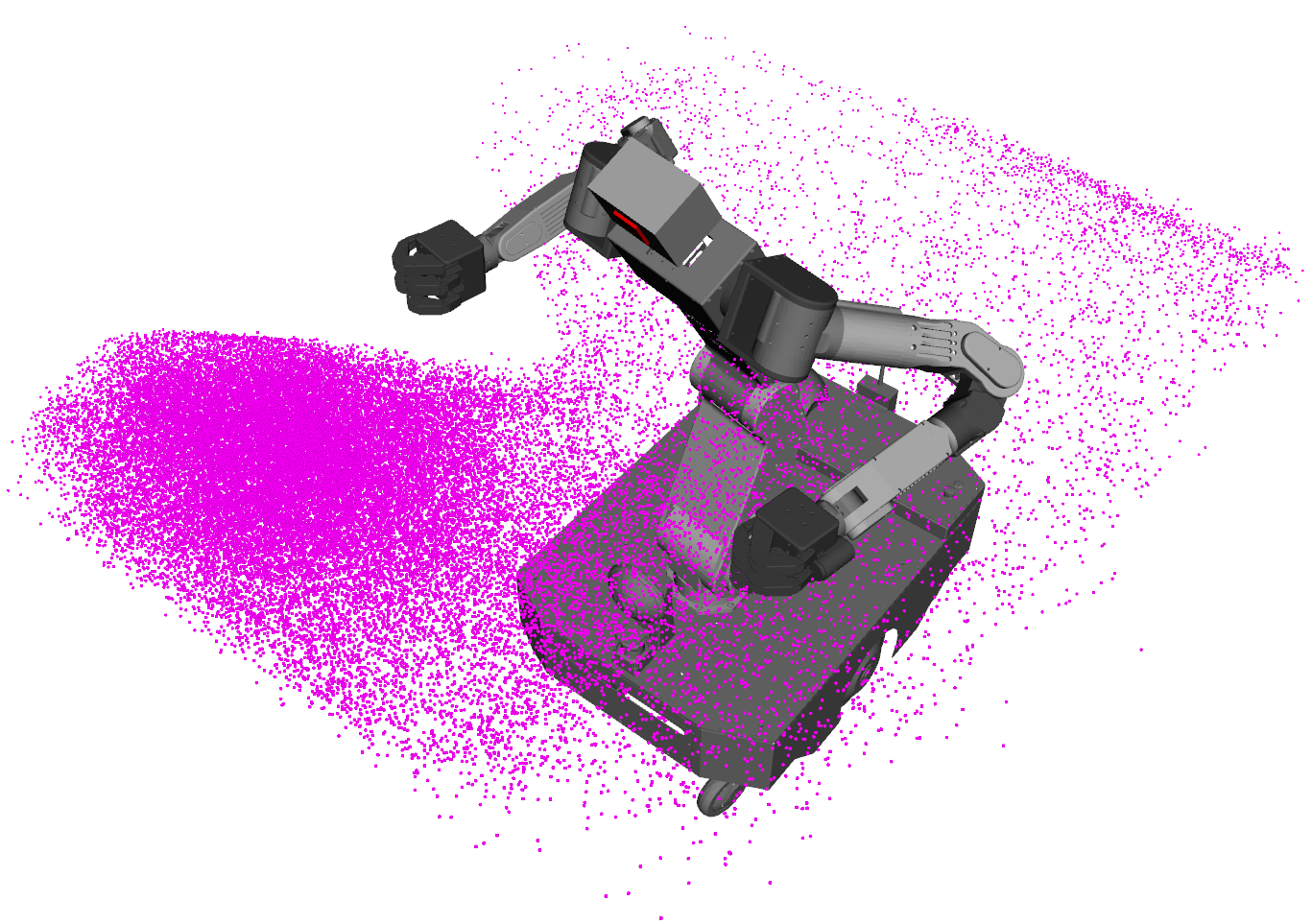}
        \caption{CGAN}
        \label{fig:manifold_app_CGAN}
    \end{subfigure}
    \caption{\small Sampling distributions for pose-constrained task using (a) CVAE and (b) CGAN. end-effector positions corresponding to constraint-satisfying samples are represented by the dots (in magenta).}
    \label{fig:manifold_app}
\end{figure}

\section{Results}
\label{sec:results}
We tested our method on different robots and environments\footnote{The reader is referred to the supplementary video accompanying this manuscript for demonstration of the results.}. Robots used include i) the \emph{Olivia} robot, a customized version of the DRC-Hubo robot comprising a 3-DOF torso and dual articulated arms (7-DOF each), and ii) the \emph{Pholus} robot, a bimanual quadrupedal robot with similar hardware and software structure to the \emph{Centauro} robot \cite{Centauro} built by the Istituto Italiano di Tecnologia (IIT). Each leg of the Pholus robot has 5 DOFs excluding the wheel, and the upper body has a single-DOF torso and a pair of 7-DOF arms. 

Our approach was implemented in MoveIt! \cite{Chitta2012} framework using the  Open Motion  Planning  Library (OMPL)~\cite{SucanOMPL}. We primarily used the RRT-Connect~\cite{Kuffner2000} planner to show the efficiency of our method, though
KPIECE (Kinodynamic Planning by Interior-Exterior Cell Exploration)~\cite{SucanKPIECE} and SBL (Single-query Bi-directional probabilistic roadmap planner)~\cite{Jaillet2010} were also tested. In addition to these planners, it is possible to use our sampler with any other available sampling-based motion planners since our method does not require any modification of existing planning algorithms.
We trained CVAE and CGAN models in Python Tensorflow using a NVIDIA GTX2080Ti GPU. Then, we used C++ implementation of Tensorflow to make inference from these models during the sampling process of planning by implementing it as a custom motion planning constraint sampler plugin in MoveIt!. 
The sampling of robot configurations from the generator of CGAN, or the decoder of CVAE, was done online during the planning.   

\subsection{End-Effector Pose-Constrained Manipulation Task}
We first compared the sampled configurations generated by CVAE and CGAN models satisfying an end-effector pose constraint where both orientation and height of the end-effector (\textbf{z} value) are to be fixed with respect to the world frame. Both CVAE and CGAN were conditioned upon the \textbf{z} value of the end-effector corresponding to different height constraint values. Using inverse kinematics, we collected 10000 robot configurations for each \textbf{z} value within a range of between 0.5-1.0m, at increments of 0.1m. 
After training, the rate of constraint-satisfaction by the samples generated from CVAE and CGAN are around 3.5\% and 9\% respectively, based on tolerance of 0.01   m for position constraint and 0.01 rad for orientation constraint. These low rates are due to the strict constraint tolerance, and the relatively small dataset size. However, it is compensated by the fact that generating samples online from CVAE and CGAN is fast, such that sufficient samples can still be generated and used. The average computation times to generate the different numbers of samples per time are shown in Table \ref{tab:sampling_time}.  To reduce the planning time, it is also possible to generate samples offline, store them in a database before planning, and then use these stored samples directly during planning. Furthermore, it is possible to generate configurations that are conditioned on height values different than the collected dataset. For instance, configurations with \textbf{z} values of 0.65, 0.77, 0.86, etc. can be sampled using the trained model, which shows the flexibility of conditional generative models.   

\begin{table}[thb]
    \caption{\small Average times for generating samples online using CVAE and CGAN models}
    \label{tab:sampling_time}
    \begin{center}
        \begin{tabular}{p{0.14\textwidth}p{0.12\textwidth}p{0.12\textwidth}}
            \toprule
            Number of Samples & CVAE (ms) & CGAN (ms)  \\
            \midrule
            100      & 3.39   &  4.03  \\
            1000      & 12.07   & 14.85  \\
            %5000      & 90.13   & 60.0 \\
            10000     & 85.6   & 119.7  \\
            %50000     & 119.7   & 619.2  \\
            100000    & 783.7   & 1214.1  \\
            \bottomrule
        \end{tabular}
    \end{center}
\end{table}

%Although samples generated by CVAE have alower rate of satisfying the constraints than CGAN, they cover the manifold more uniformly compared to CGAN.
Although samples generated by CVAE have a lower rate of satisfying the constraints than CGAN, they cover the manifold more uniformly compared to CGAN. While the samples generated by CGAN model have a higher rate of constraint satisfaction, they cover specific regions of the manifold more, which is a well-known issue known as mode collapse for GANs  \cite{Salimans2016}. More advanced GAN models in the literature  address this issue \cite{Arjovsky2017}, but it is out of the scope of this paper. For the same amount of sampled configurations, the distribution of the constraint-satisfying samples can be seen in Figure \ref{fig:manifold_app}, where each dot represents the end-effector position corresponding to a valid constraint-satisfying robot configuration generated by the model.

\begin{figure*}[thb]
    \centering
    \begin{subfigure}{0.15\textwidth}
        \includegraphics[width=\textwidth]{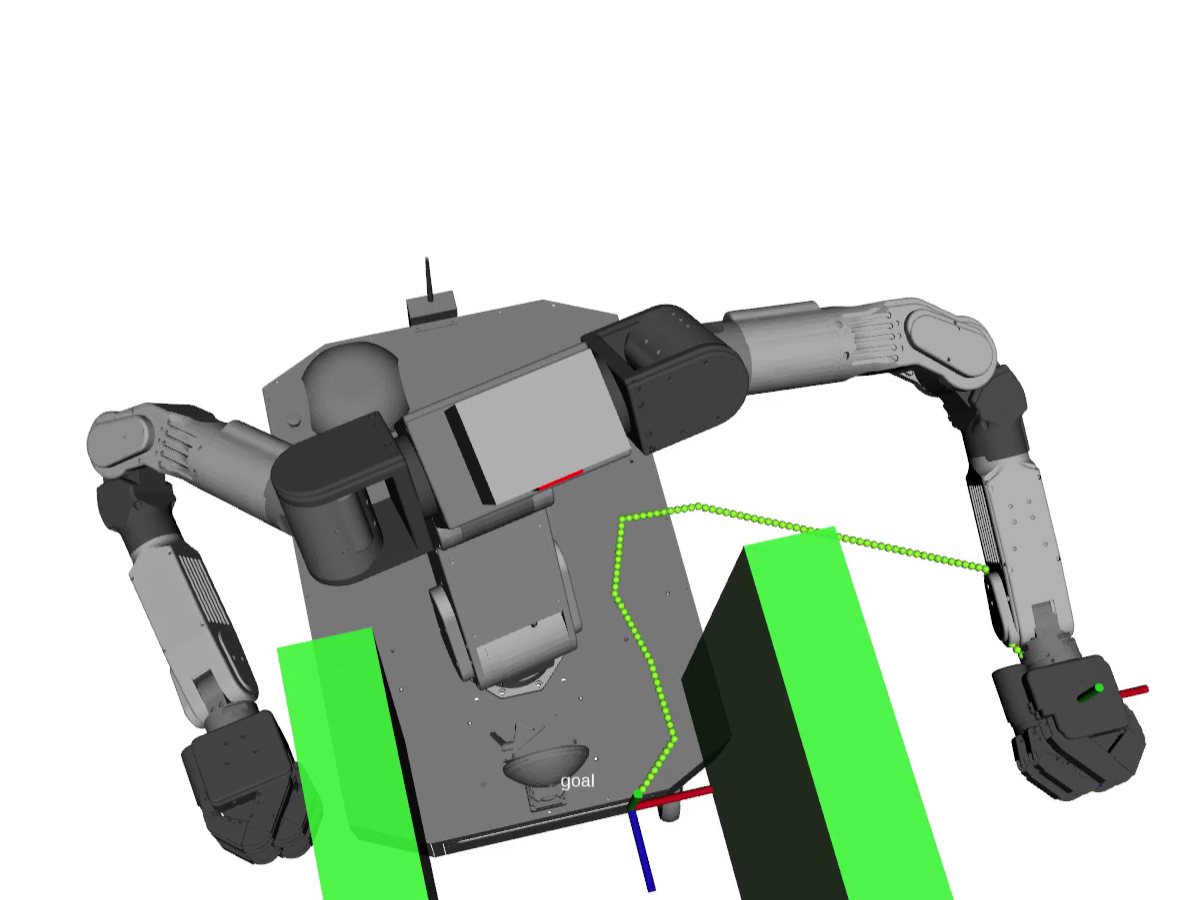} 
        \caption{$t_0$ (initial)}
        \label{fig:wall0}
    \end{subfigure}
    \begin{subfigure}{0.15\textwidth}
        \includegraphics[width=\textwidth]{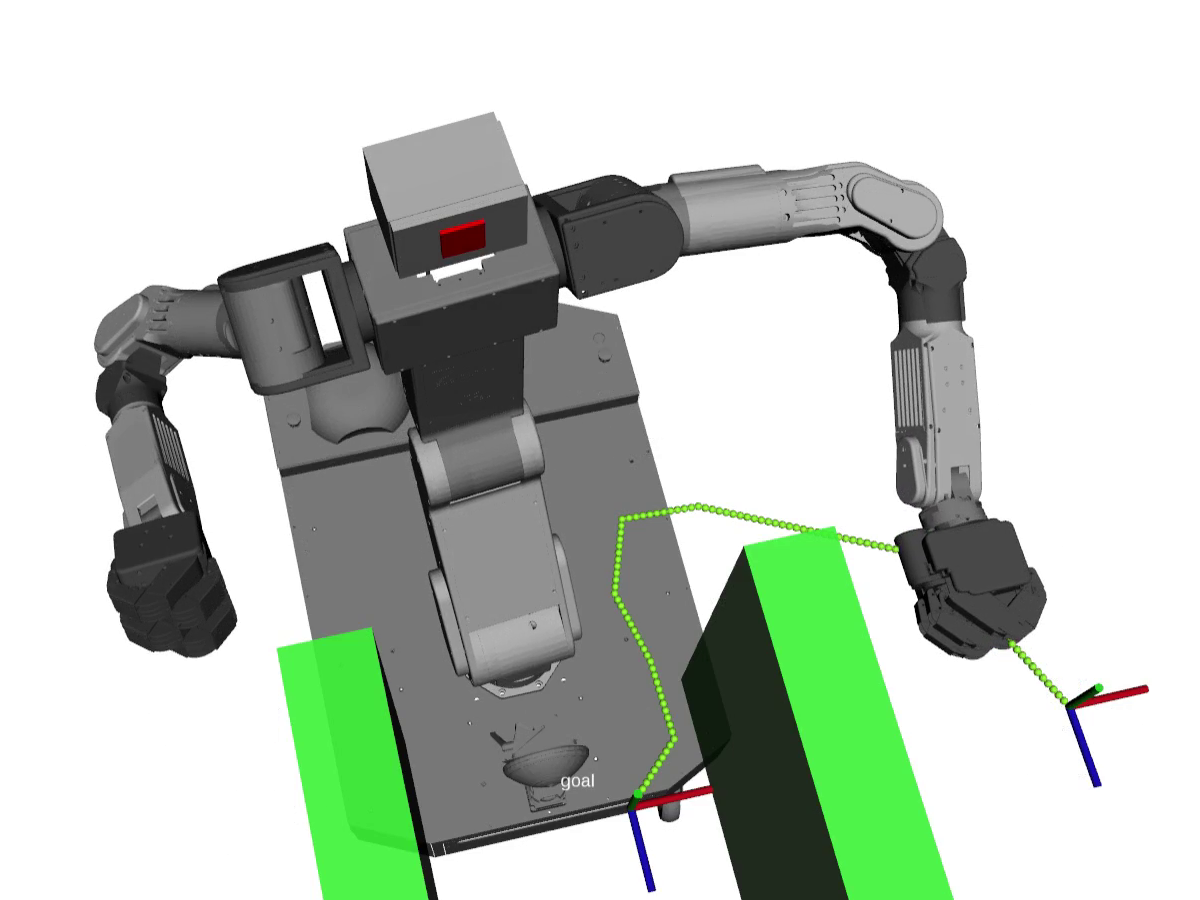} 
        \caption{$t_1$}
        \label{fig:wall1}
    \end{subfigure}
    \begin{subfigure}{0.15\textwidth}
        \includegraphics[width=\textwidth]{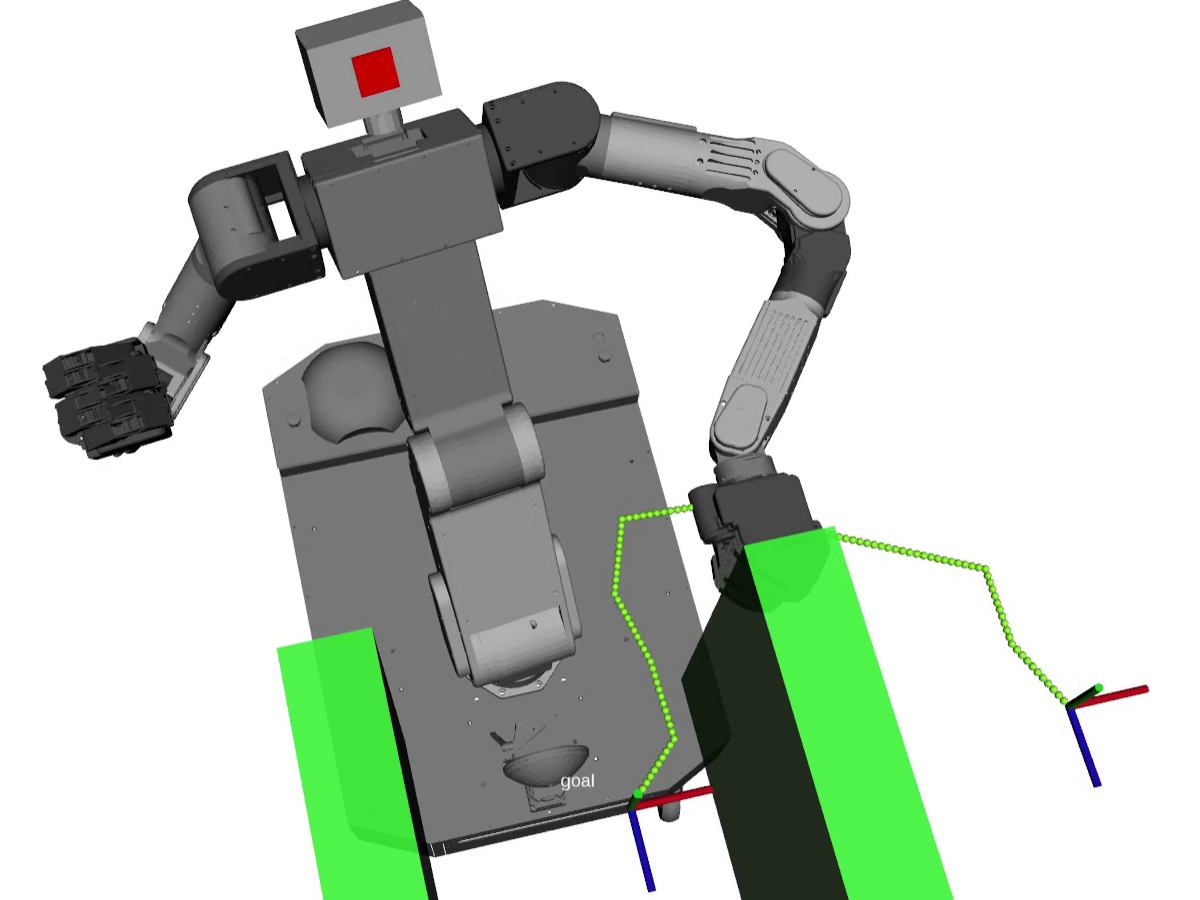}  
        \caption{$t_2$}
        \label{fig:wall2}
    \end{subfigure}
    \begin{subfigure}{0.15\textwidth}
        \includegraphics[width=\textwidth]{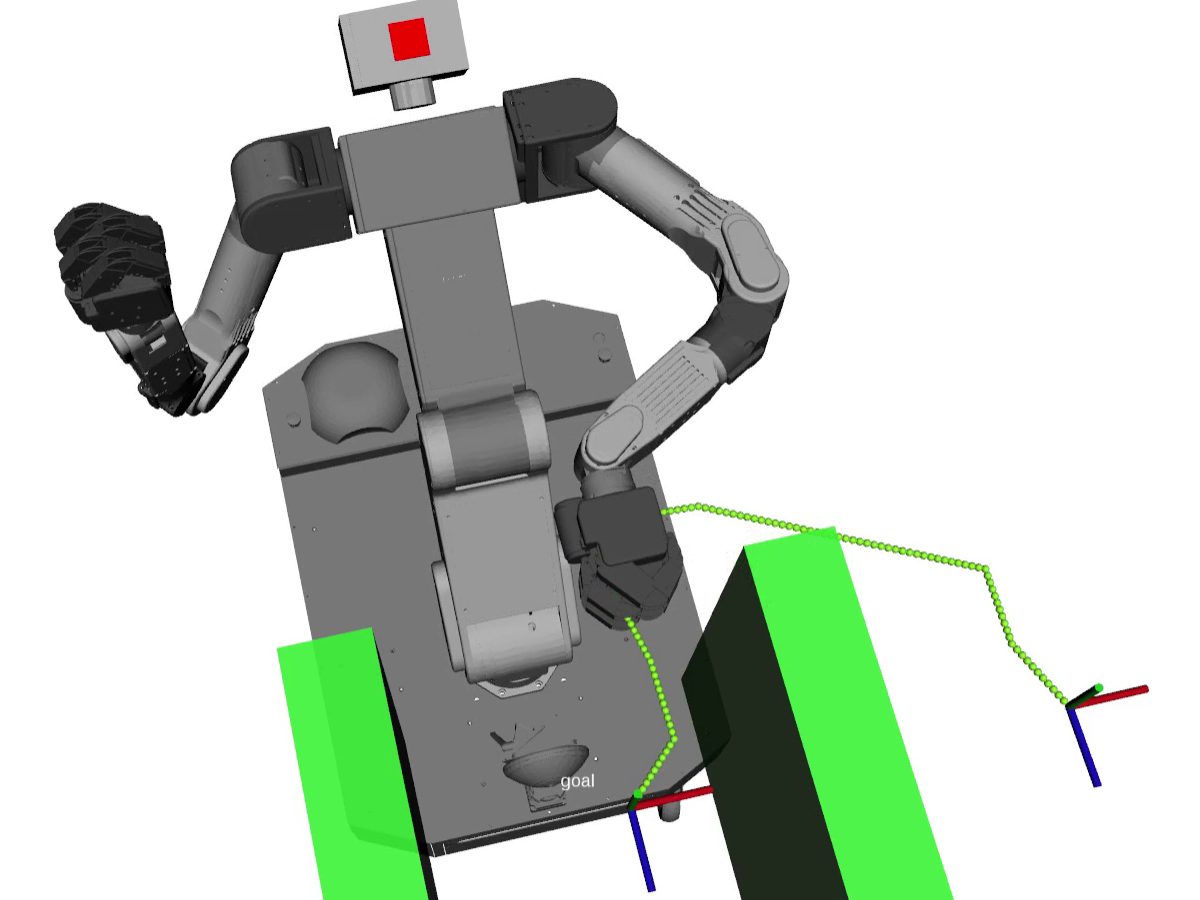}  
        \caption{$t_3$}
        \label{fig:wall3}
    \end{subfigure}
    \begin{subfigure}{0.15\textwidth}
        \includegraphics[width=\textwidth]{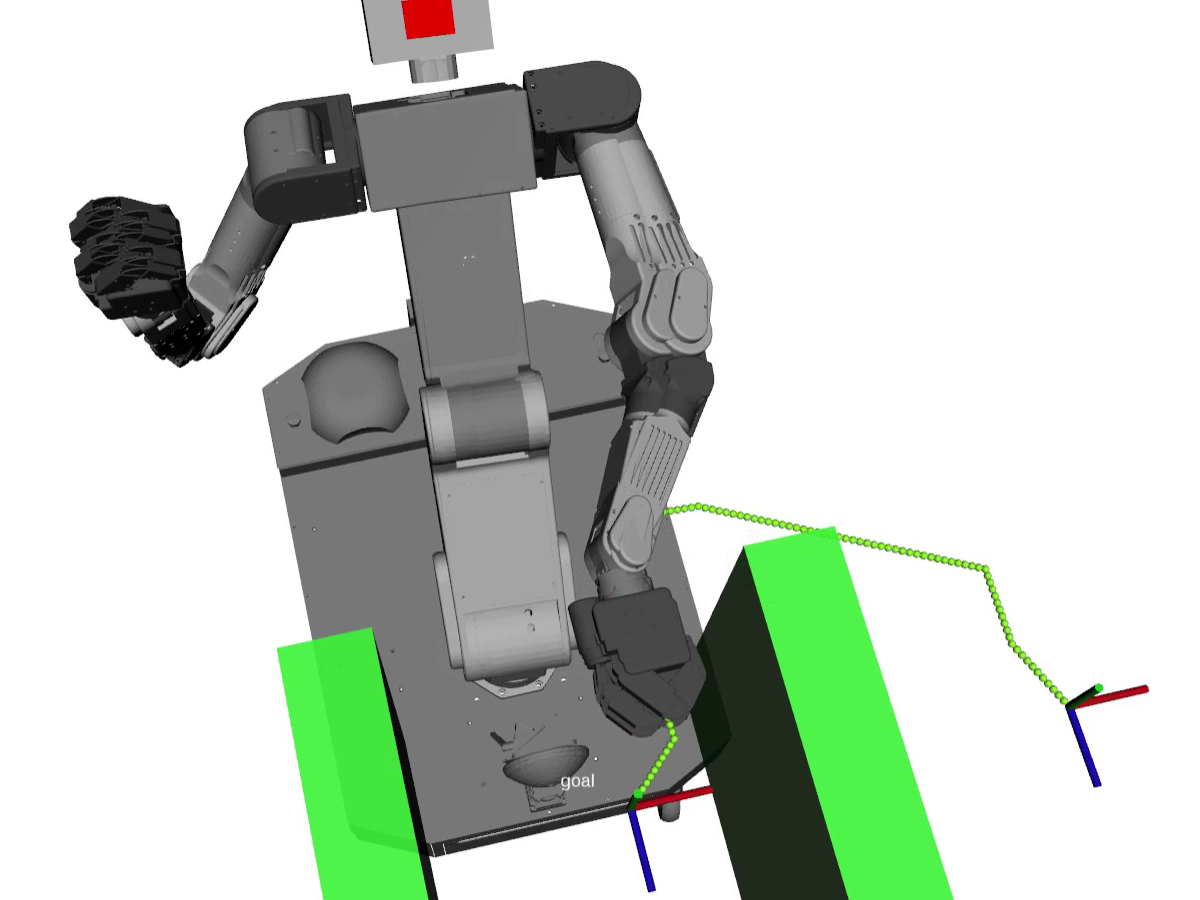}  
        \caption{$t_4$}
        \label{fig:wall4}
    \end{subfigure}
    \begin{subfigure}{0.15\textwidth}
        \includegraphics[width=\textwidth]{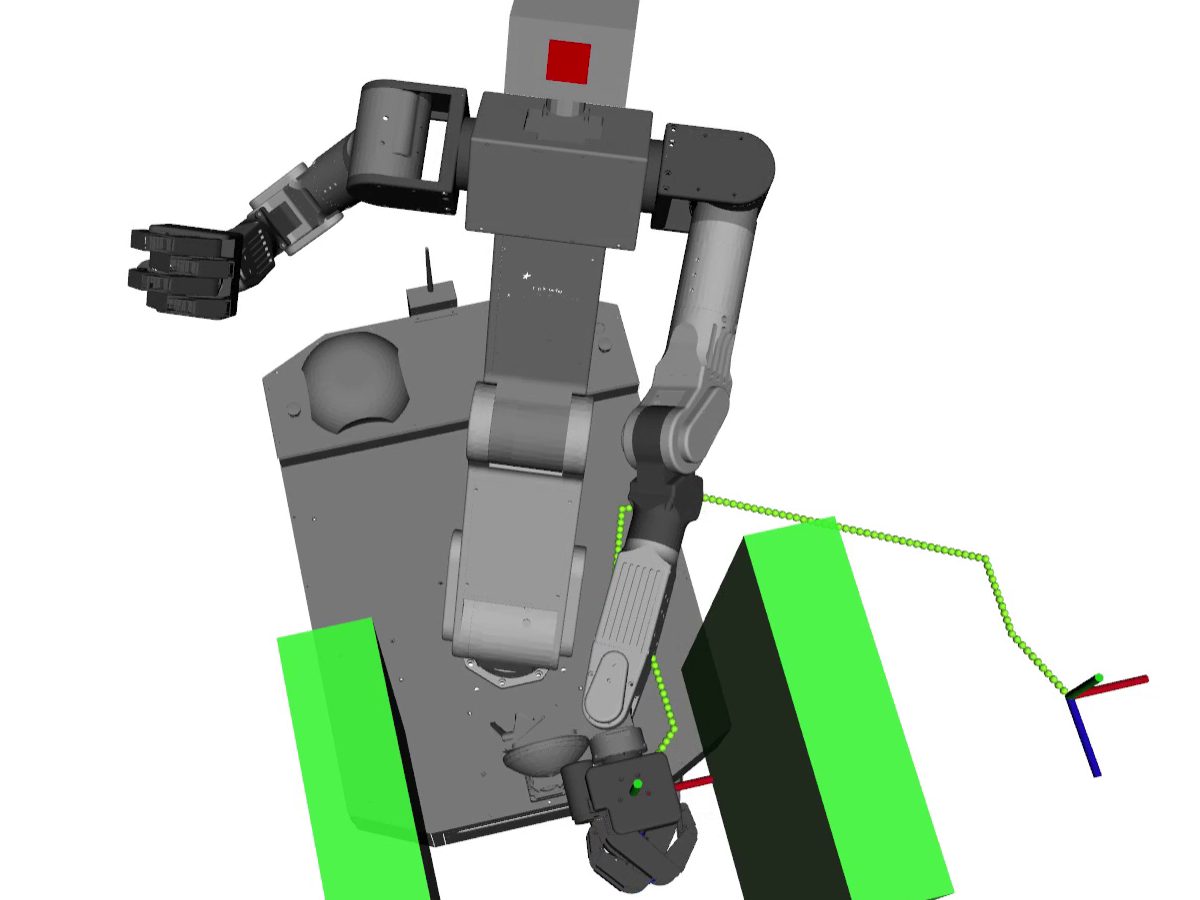}  
        \caption{$t_5$ (goal)}
        \label{fig:wall5}
    \end{subfigure}
    \caption{\small Pose-constrained manipulation task where the end-effector needs to move from start to goal (marked with axes) on both sides of a wall, subject to constant height and orientation. The solution path obtained using CGAN model is illustrated with green dot. }
    \label{fig:wall_CGAN}
\end{figure*}

For evaluation of the generative models, we first considered a relatively simple environment with obstacles as seen in Figure \ref{fig:wall_CGAN}. The joints in the left arm and torso (total 10 DOFs) were used for planning, under the constraint of fixed orientation in roll, pitch, yaw axes, and height of the end-effector. The goal location was between two wall-shaped obstacles, and the start location on one side. Besides CVAE and CGAN, we also evaluated, in this environment, a projection-based constrained motion planning method, namely CBiRRT \cite{Berenson2011}, and a method of approximating constraint manifolds with precomputed graph (PG)  \cite{Sucan2012}. The PG dataset consisted of the same 10000 configurations used to train the generative models. The success rate and average planning time without path simplification are shown in Table \ref{tab:wall_problem} for 1000 runs. We observe that CVAE and CGAN have both faster planning times on average than CBiRRT.      

\begin{figure*}[thb]
    \centering
    \begin{subfigure}{0.15\textwidth}
        \includegraphics[width=\textwidth]{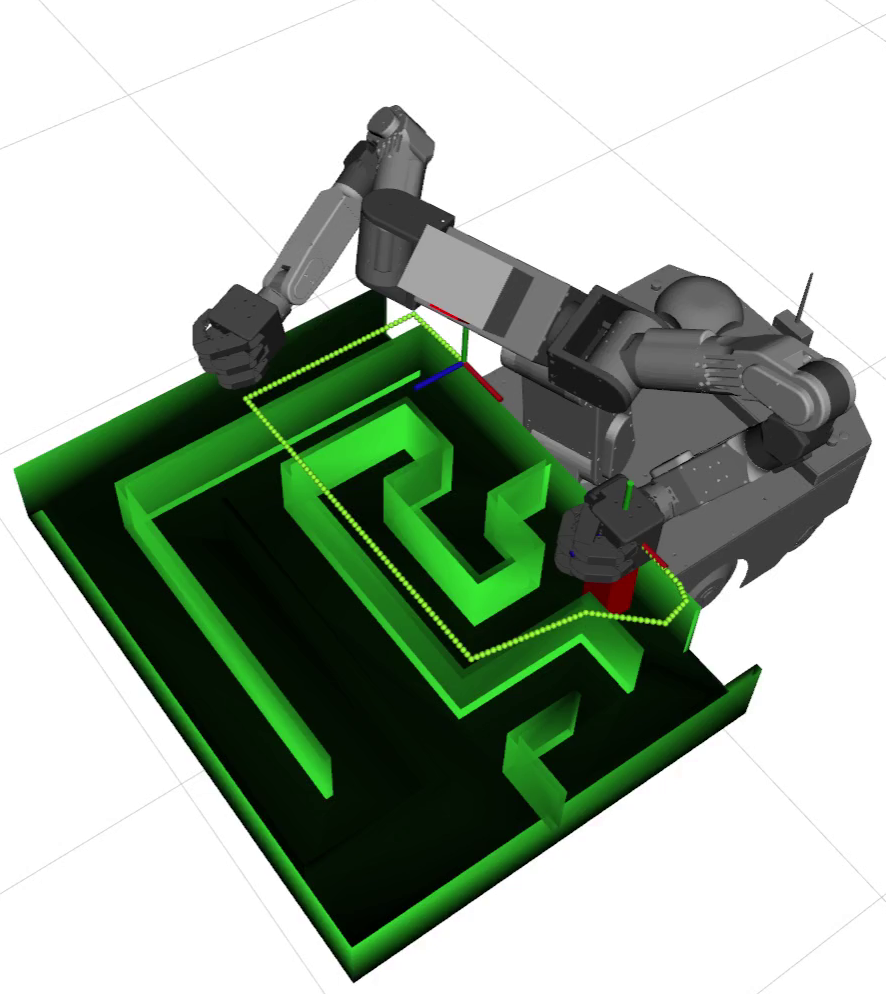} 
        \caption{$t_0$ (initial)}
        \label{fig:maze0}
    \end{subfigure}
    \begin{subfigure}{0.15\textwidth}
        \includegraphics[width=\textwidth]{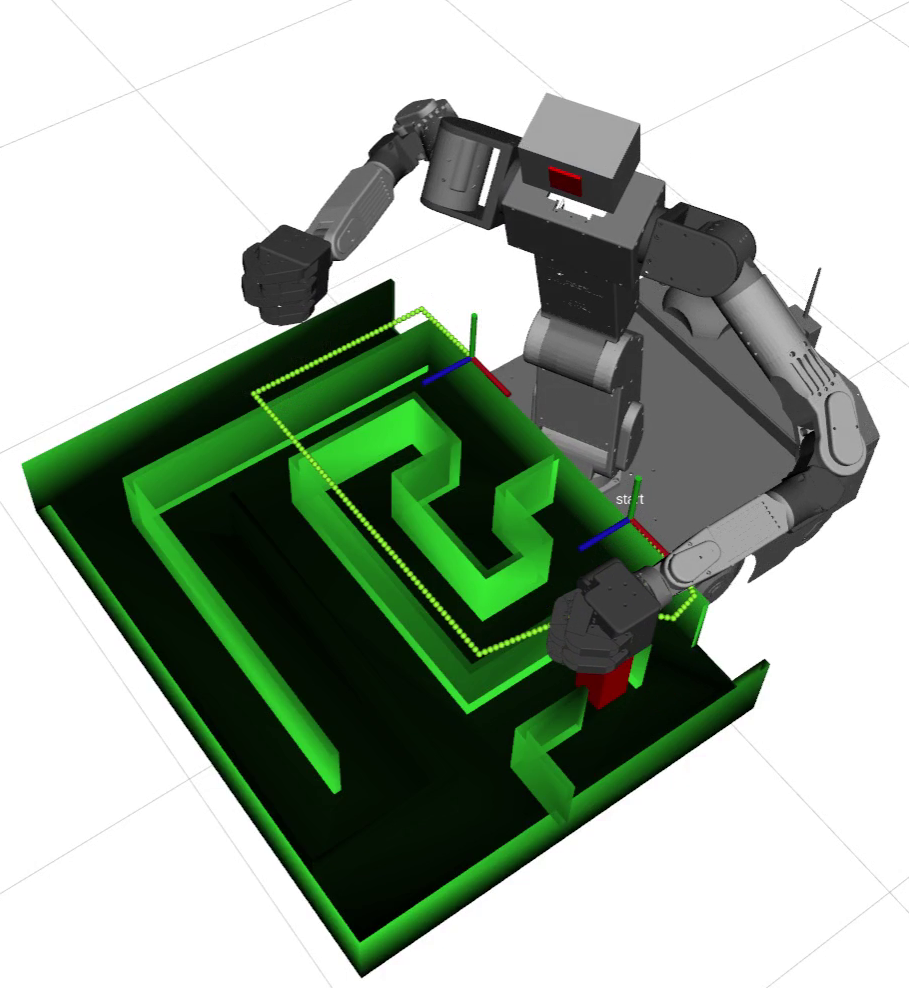} 
        \caption{$t_1$}
        \label{fig:maze1}
    \end{subfigure}
    \begin{subfigure}{0.15\textwidth}
        \includegraphics[width=\textwidth]{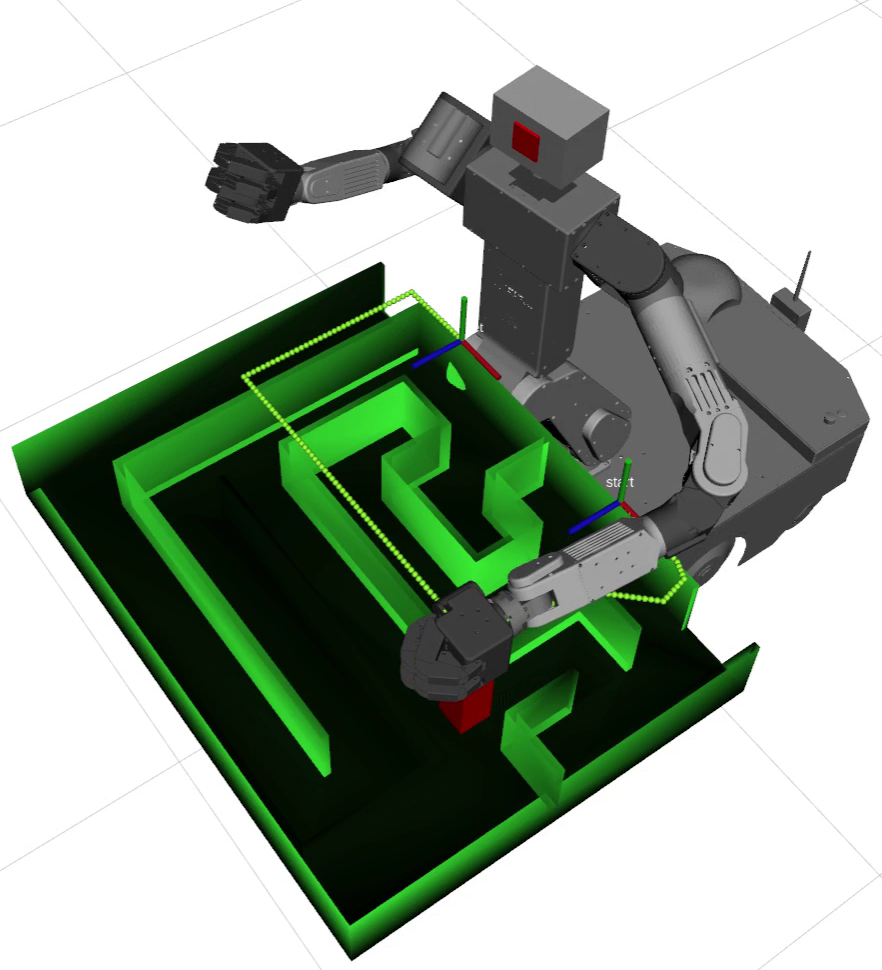}  
        \caption{$t_2$}
        \label{fig:maze2}
    \end{subfigure}
    \begin{subfigure}{0.15\textwidth}
        \includegraphics[width=\textwidth]{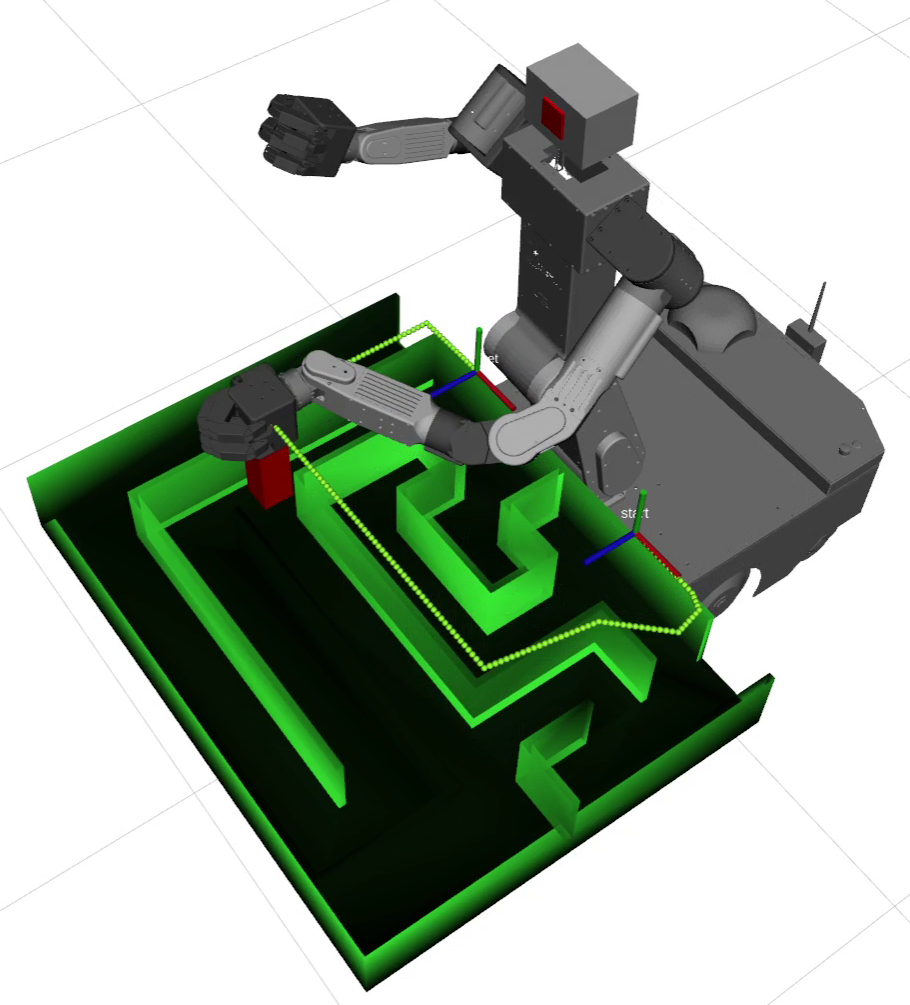}  
        \caption{$t_3$}
        \label{fig:maze3}
    \end{subfigure}
    \begin{subfigure}{0.15\textwidth}
        \includegraphics[width=\textwidth]{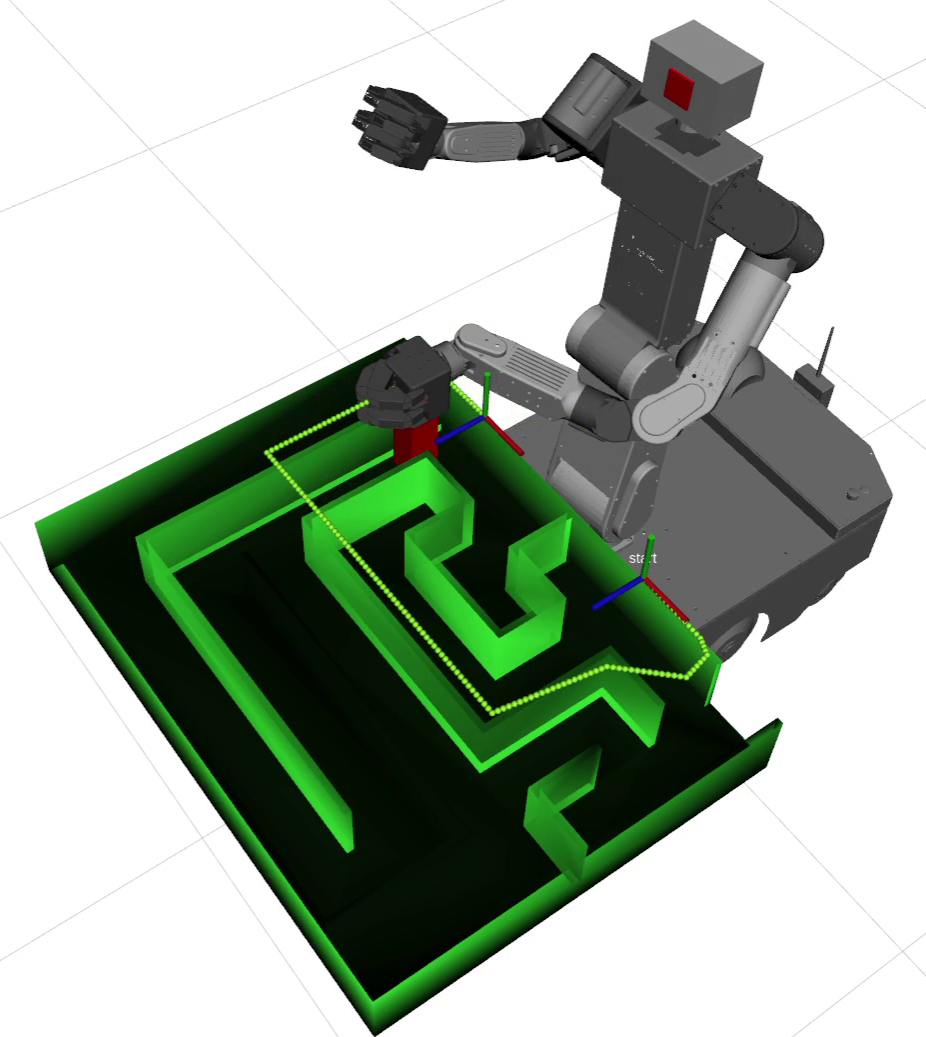}  
        \caption{$t_4$}
        \label{fig:maze4}
    \end{subfigure}
    \begin{subfigure}{0.15\textwidth}
        \includegraphics[width=\textwidth]{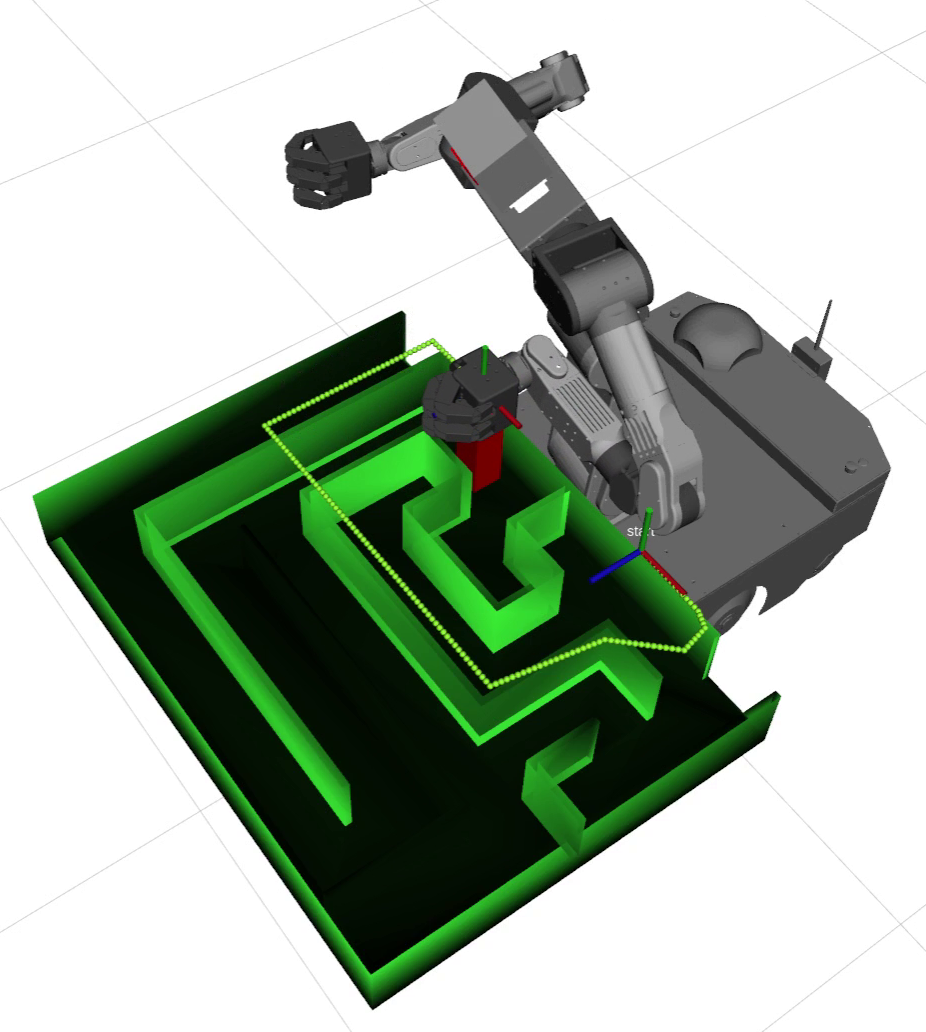}  
        \caption{$t_5$ (goal)}
        \label{fig:maze5}
    \end{subfigure}
    \caption{\small Maze problem scene. End-effector of Olivia holding a tool is required to keep the height constant. The start and goal positions are marked with axes. The solution path obtained using CVAE model is illustrated with green dots. }
    \label{fig:maze_CVAE}
\end{figure*}

\begin{figure*}[h]
    \centering
    \begin{subfigure}{0.15\textwidth}
        \includegraphics[width=\textwidth]{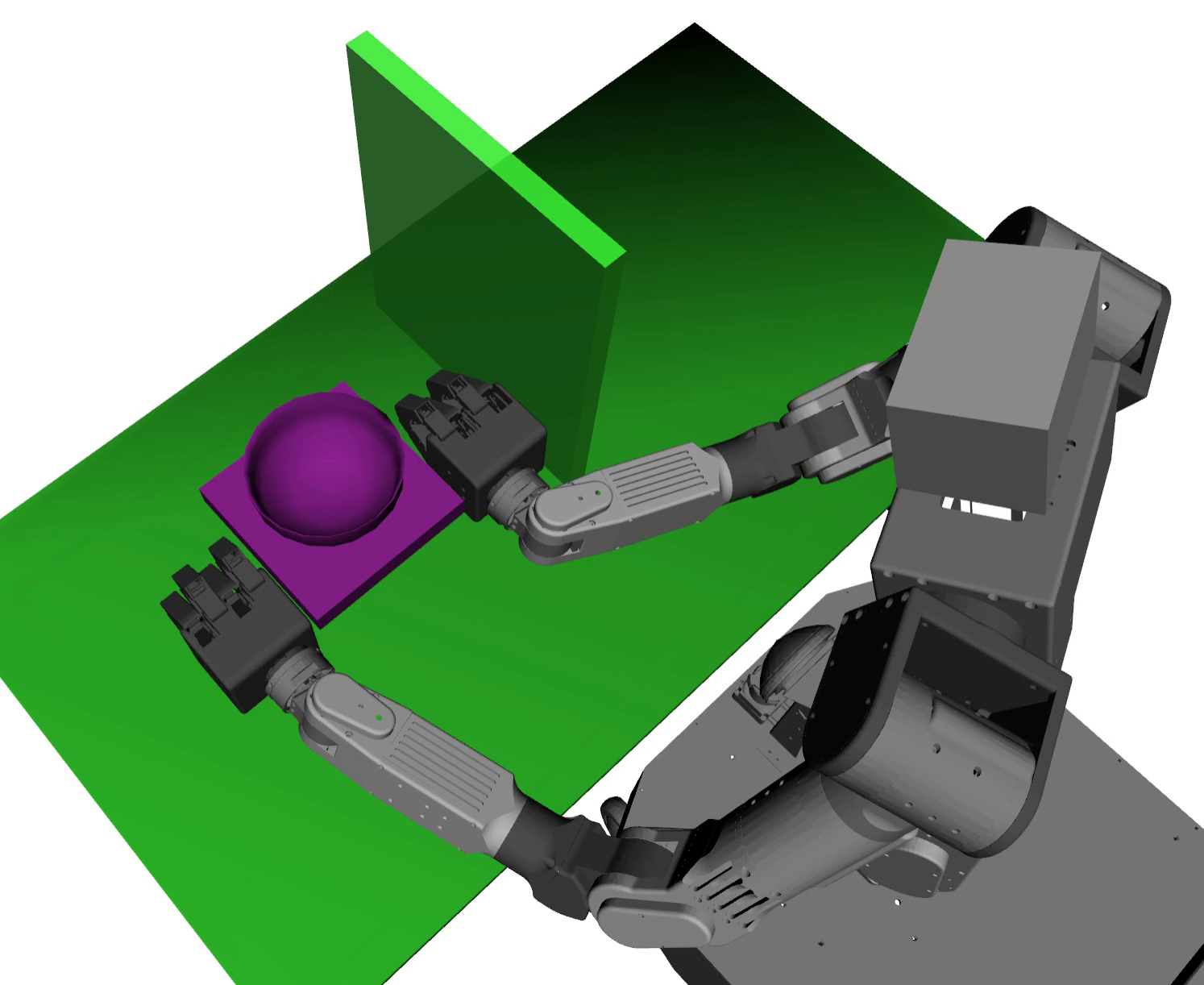} 
        \caption{$t_0$ (initial)}
        \label{fig:dual0}
    \end{subfigure}
    \begin{subfigure}{0.15\textwidth}
        \includegraphics[width=\textwidth]{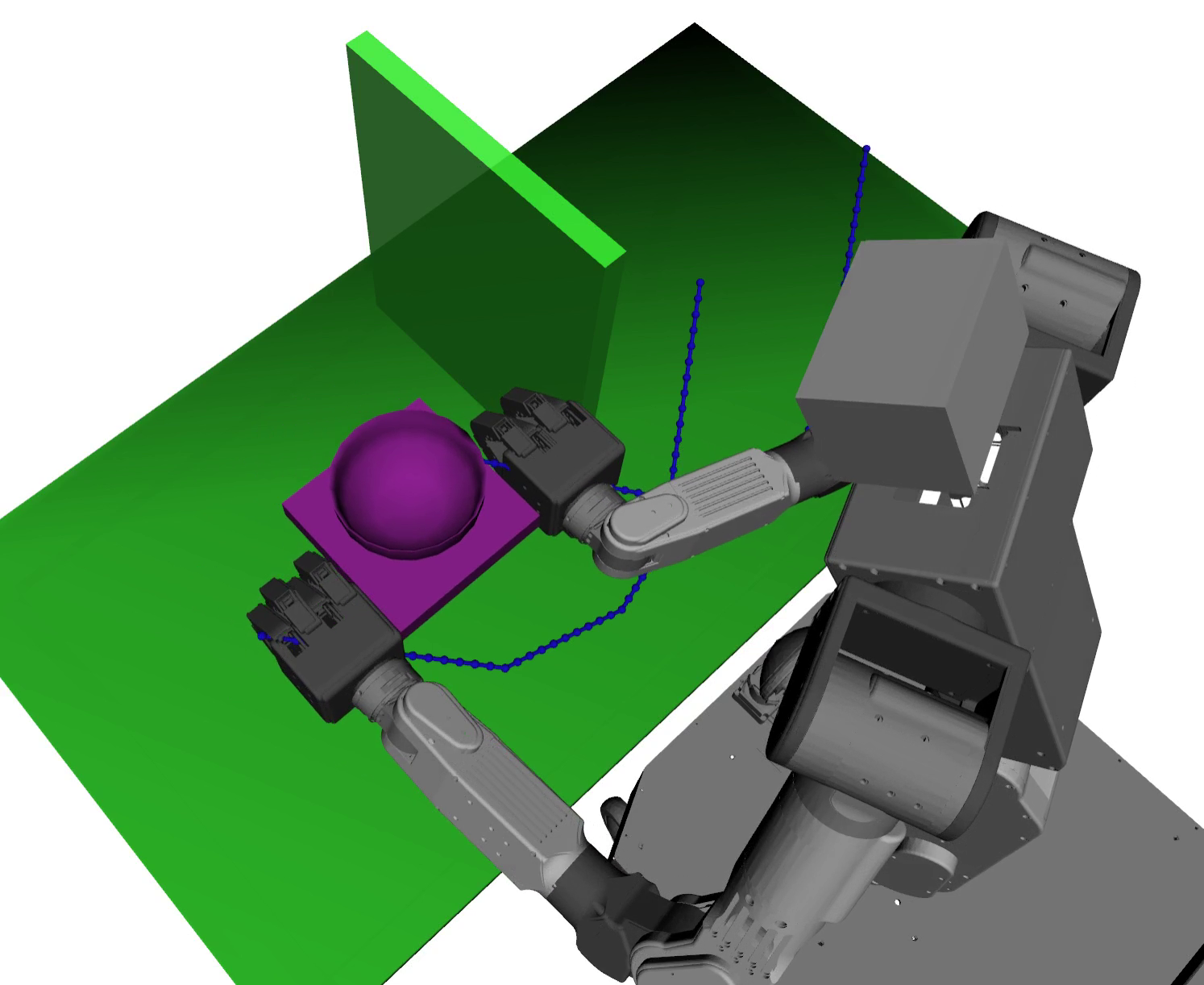} 
        \caption{$t_1$}
        \label{fig:dual1}
    \end{subfigure}
    \begin{subfigure}{0.15\textwidth}
        \includegraphics[width=\textwidth]{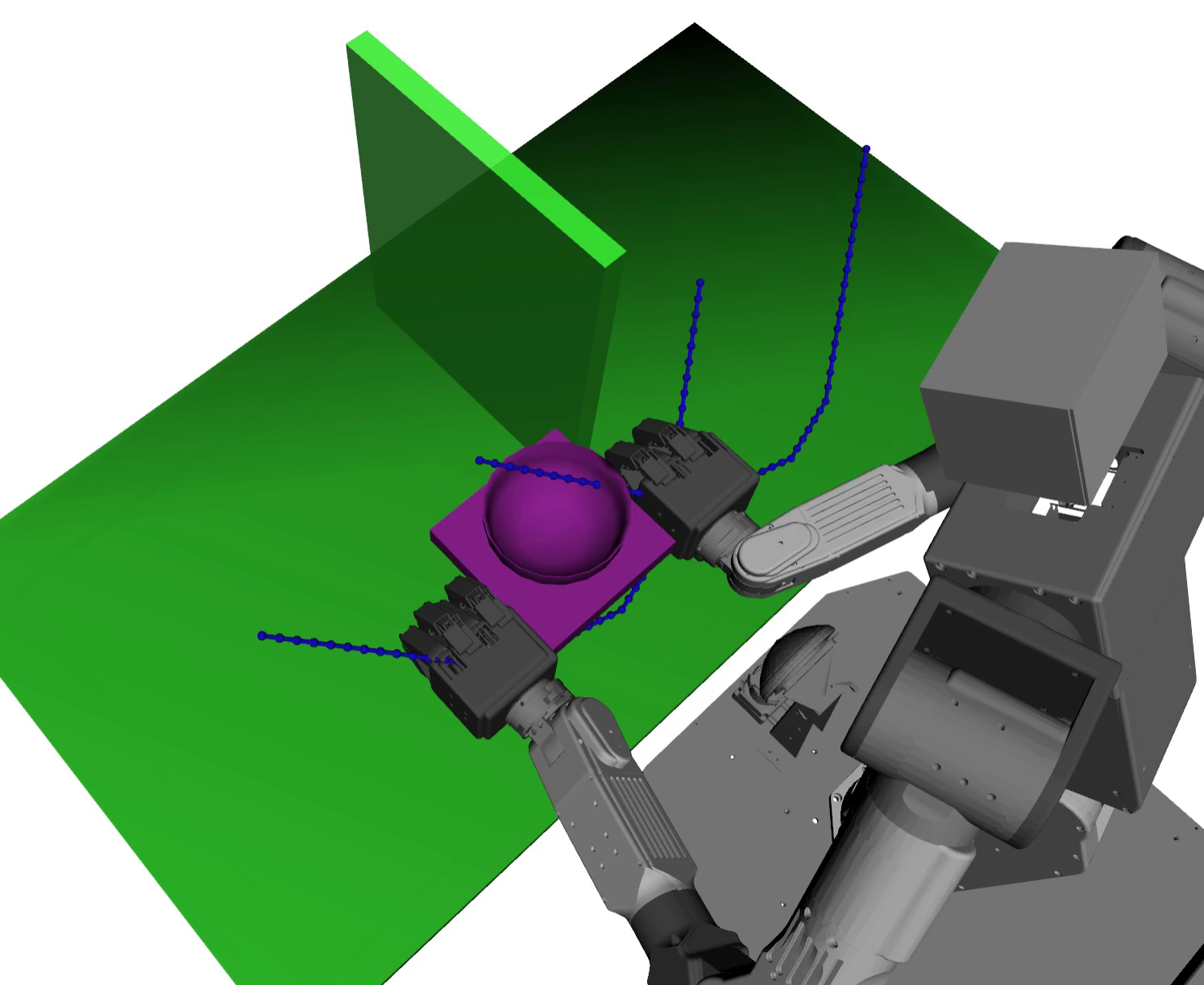} 
        \caption{$t_2$}
        \label{fig:dual2}
    \end{subfigure}
    \begin{subfigure}{0.15\textwidth}
        \includegraphics[width=\textwidth]{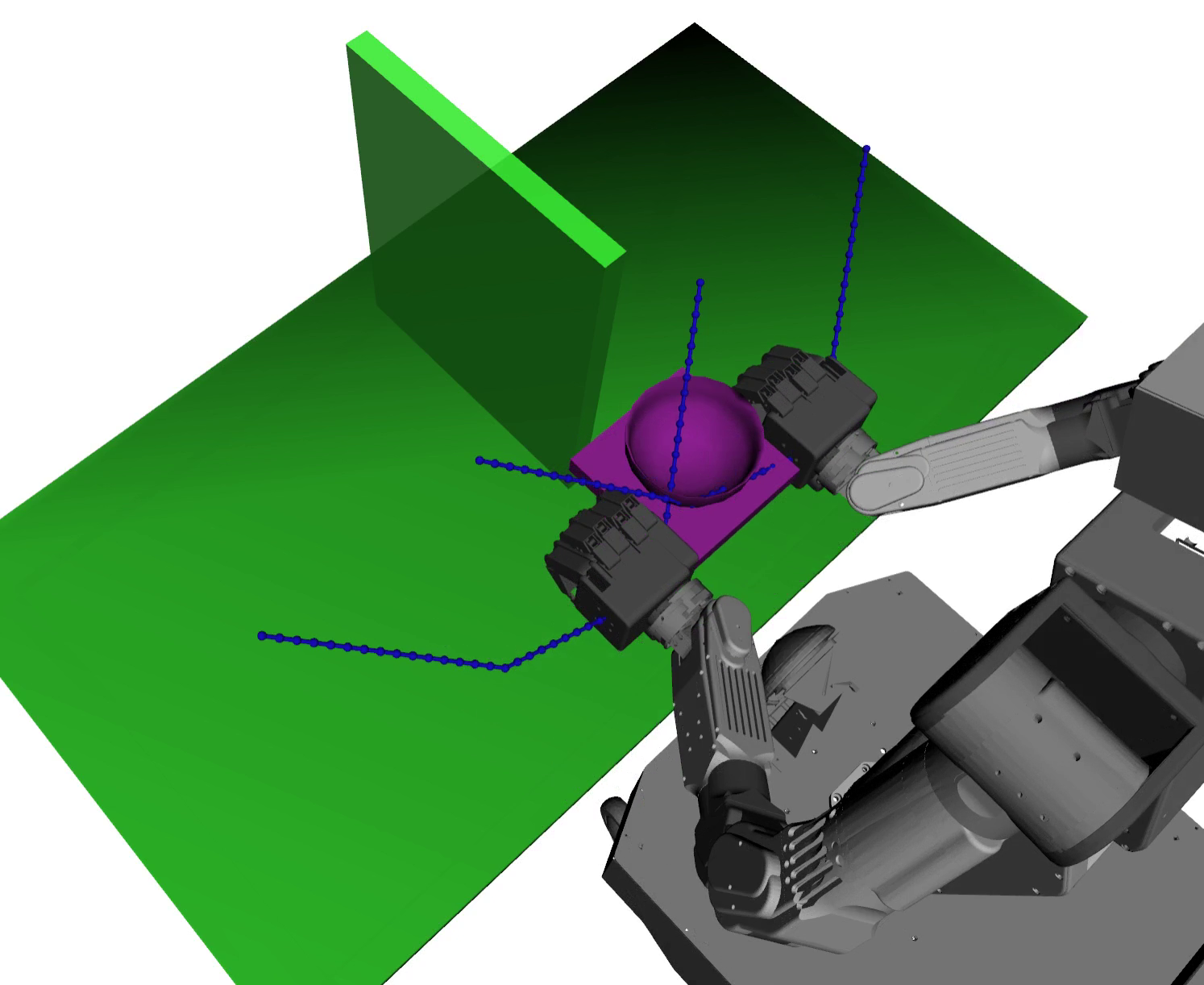} 
        \caption{$t_3$}
        \label{fig:dual3}
    \end{subfigure}
    \begin{subfigure}{0.15\textwidth}
        \includegraphics[width=\textwidth]{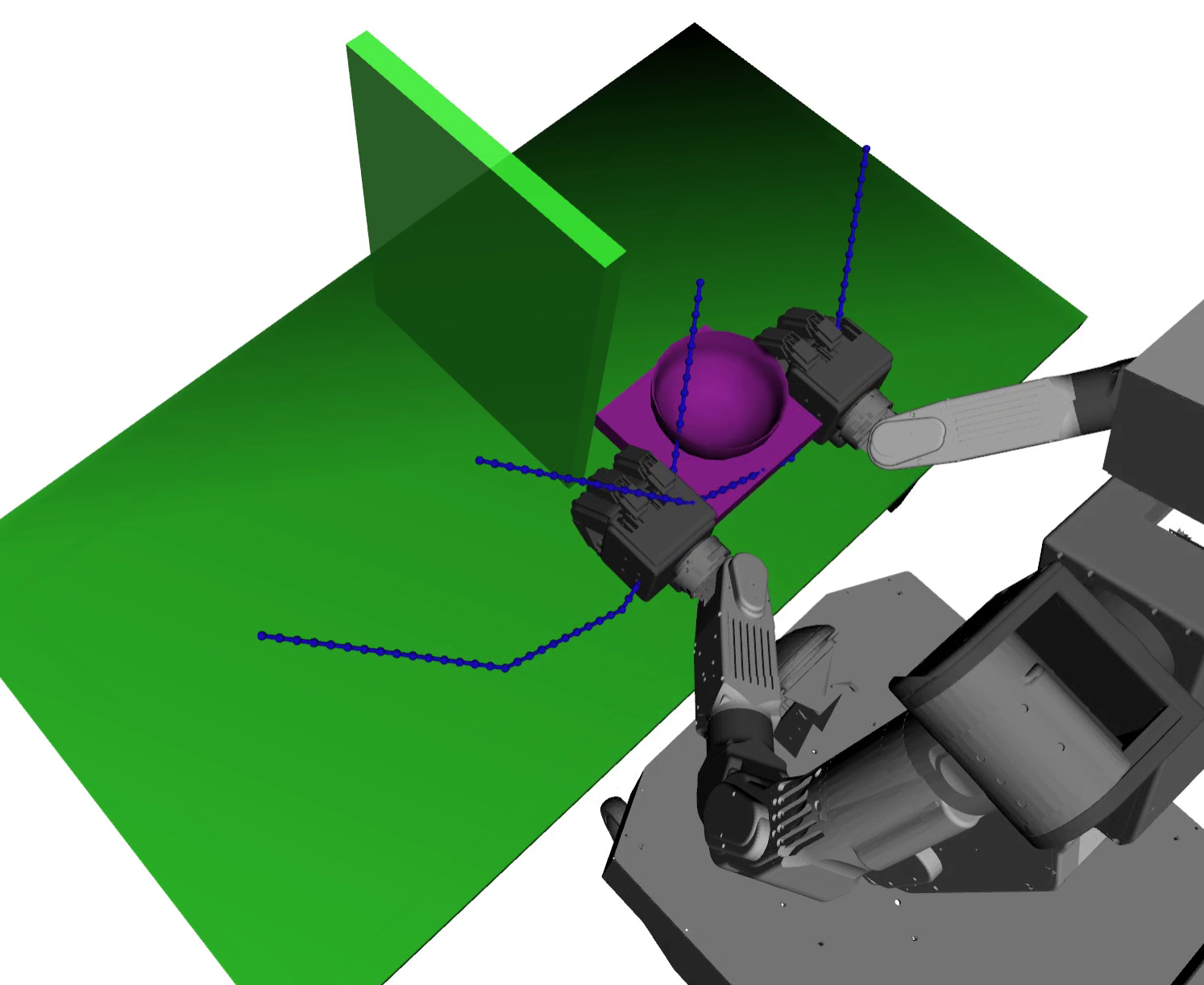} 
        \caption{$t_4$}
        \label{fig:dual4}
    \end{subfigure}
    \begin{subfigure}{0.15\textwidth}
        \includegraphics[width=\textwidth]{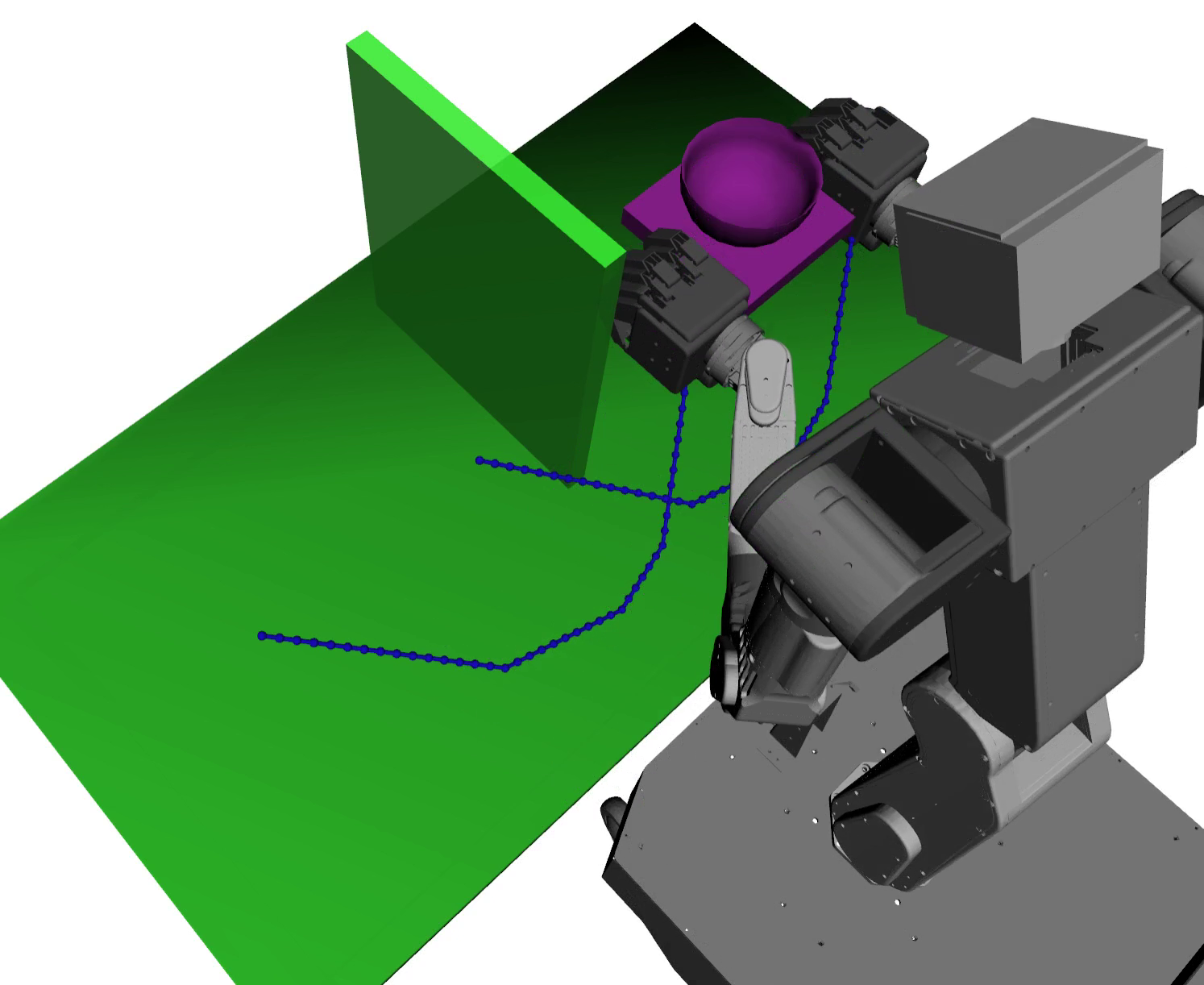} 
        \caption{$t_5$ (goal)}
        \label{fig:dual5}
    \end{subfigure}
    \caption{\small Snapshots of dual-arm manipulation of object to goal pose in the presence of object orientation constraint and obstacle.}
    \label{fig:dual-arm}
\end{figure*}

\begin{table}[h]
    \caption{\small Computation times and success rates for 1000 runs of the obstacle avoidance problem.}
    \label{tab:wall_problem}
    \begin{center}
        \begin{tabular}{p{0.07\textwidth}p{0.07\textwidth}p{0.07\textwidth}p{0.07\textwidth}p{0.07\textwidth}}
            \toprule
            Methods & Average Time(s) & Minimum Time(s) & Maximum Time(s) & Success Rate (\%) \\
            \midrule
            CVAE  &  0.116   & 0.085   & 0.32    & 100.0  \\
            CGAN  &  0.164   & 0.105  &  0.42    & 100.0  \\
            PG  &  0.077   & 0.039  &  0.3    &   97.6  \\
            CBiRRT   & 0.572   & 0.076  & 2.8    & 100.0  \\
            \bottomrule
        \end{tabular}
    \end{center}
\end{table}

Next, we tested the efficiency of these models in a more difficult maze environment, as seen in Figure \ref{fig:maze_CVAE}, under the same end-effector pose constraints. Even though the environment obstacles were different, the same trained CVAE and CGAN models of the pose constraint manifold were used. The robot, holding a tool (red cuboid object), must plan motion from a start configuration to a goal configuration while keeping the height of the tool constant with respect to the world frame, and also the roll, pitch, and yaw axes fixed within a tolerance of 0.01 rad. Even though the task constraints are the same as in the previous task, this is a much harder problem to solve since collision avoidance in a maze environment causes a significant reduction in the dimension of the constraint manifold. 
The maximum time to find a solution for motion planning is limited to 200 seconds. After 200 seconds, motion planning is terminated and counted as a failure.
Our CBiRRT implementation did not manage to find a solution for the given time limit. The results in Table \ref{tab:maze_problem}, based on 100 runs, show that the highest success rate and shortest planning time came from the CVAE generative model, which indicates the importance of covering sampling distribution uniformly. We see that since sampling from a precomputed dataset is limited to a fixed number of configurations, reduction in the dimension of the constraint manifold results in a notable performance drop for PG in this maze environment. A visualization of the motion plan found using the CVAE model can be seen in Figure \ref{fig:maze_CVAE}, which shows the end-effector path and also snapshots of the robot joint configurations.

\begin{table}[h]
    \caption{\small Computation times and success rates for 100 runs of the maze problem.}
    \label{tab:maze_problem}
    \begin{center}
        \begin{tabular}{p{0.07\textwidth}p{0.07\textwidth}p{0.07\textwidth}p{0.07\textwidth}p{0.07\textwidth}}
            \toprule
            Methods & Average Time(s) & Minimum Time(s) & Maximum Time(s) & Success Rate (\%) \\
            \midrule
            CVAE  & 18.83   & 6.15   & 46.06     & 99.0  \\
            CGAN  & 93.81  & 16.81  & 193.87    &  89.0  \\
            PG   & 119.7   & 35.18  & 200.02    & 37.0  \\
            \bottomrule
        \end{tabular}
    \end{center}
\end{table}

\subsection{ Dual-Arm-Constrained Manipulation Task}
To evaluate our method in handling closed-loop kinematic constraints, we considered a task with the Olivia robot bimanually holding a tray with a bowl on top. This task involved both orientation and position constraints for one arm with respect to the other, in order to hold the tray and also an orientation constraint for both arms with respect to the world frame to keep the bowl upright. As a result, roll, pitch and yaw values of the tray were to be kept fixed. CVAE and CGAN were conditioned on the parameter $d$ representing the distance between the wrists and also orientation of the wrists, so as to hold an object at different poses. We collected robot configurations for a set of $d$ values within a range of between 0.2-0.7m, at increments of 0.1m. For each $d$ value, we collected 2500 configurations for each of 4 different pitch angles of the wrist (at 45$^{\circ}$ intervals), giving a total of 10000 configurations per $d$ value. In addition, two 7-DOF arms, together with a 3-DOF torso, make up a 17-dimensional configuration space. We collected dual-arm pose-constrained configurations using the Bio-IK \cite{Ruppel2018} algorithm (as a MoveIt! plugin) to deal with multi-goal end-effector tasks. To implement dual-arm manipulation planning, MoveIt! source code was also modified. The execution of a representative motion plan found by using CVAE, with RRT-Connect planner, $d=0.3$, and maximum planning time of 200s, can be seen in Figure \ref{fig:dual-arm}. Results for 100 runs with different sampling-based planners, under default settings, can be seen in Table \ref{tab:dual_arm}. As seen, CVAE outperforms CGAN across all different planning algorithms. We observe that mode collapse for CGAN is the main reason for this result. In terms of planners, RRT and RRTConnect planners perform better compared to KPIECE and SBL planners, which is consistent with results in \cite{Sucan2012}.

\begin{table}[ht]
    \caption{\small Computation times and success rates for the dual-arm constraint problem with different sampling-based motion planners}
    \label{tab:dual_arm}
    \begin{center}
        \begin{tabular}{p{0.07\textwidth}p{0.07\textwidth}p{0.07\textwidth}p{0.07\textwidth}p{0.07\textwidth}}
            \toprule
            \textbf{Methods} & \multicolumn{2}{c}{\textbf{CVAE}} & \multicolumn{2}{c}{\textbf{CGAN}} \\
            \cmidrule{2-5}
            & Average Time(s) & Success Rate (\%) &Average Time(s) & Success Rate (\%) \\
            \midrule
            RRT         & 9.64      & 98.0              & 80.4     & 50.0  \\
            RRTConnect  & 9.88      & 99.0              & 61.05     & 61.0  \\
            KPIECE      & 12.14     & 100.0              & 103.24    & 41.0  \\
            SBL         & 105.84    & 81.0               & 140.88    & 23.0  \\
            \bottomrule
        \end{tabular}
    \end{center}
\end{table}

\subsection{ Balance-Constrained Whole-Body Motion Task}
\begin{figure}[h]
    \centering
    \begin{subfigure}{0.13\textwidth}
        \includegraphics[width=\textwidth]{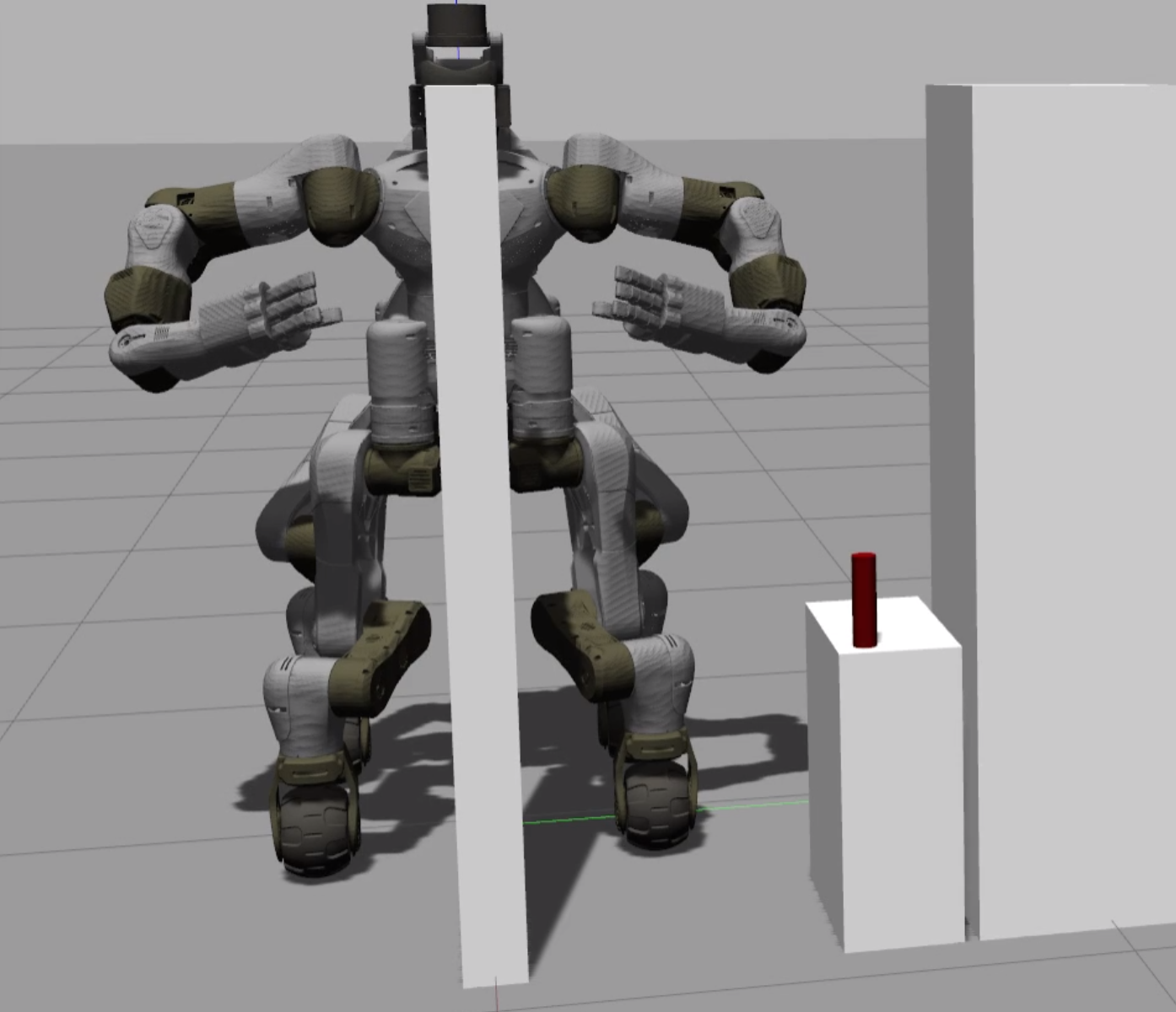} 
        \caption{$t_0$ (initial)}
        \label{fig:subim1}
    \end{subfigure}
    \begin{subfigure}{0.13\textwidth}
        \includegraphics[width=\textwidth]{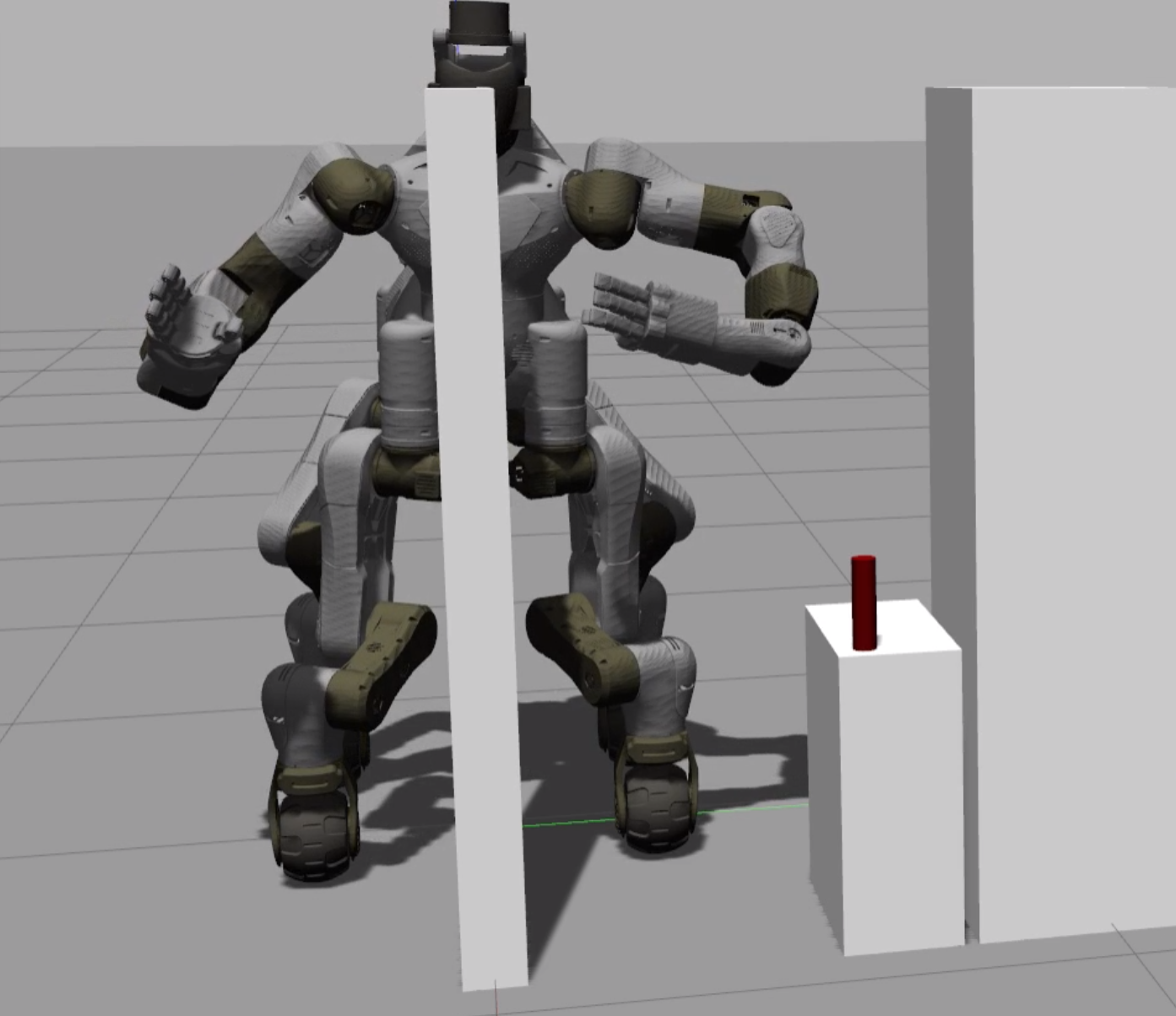}
        \caption{$t_1$}
        \label{fig:subim3}
    \end{subfigure}
    \begin{subfigure}{0.13\textwidth}
        \includegraphics[width=\textwidth]{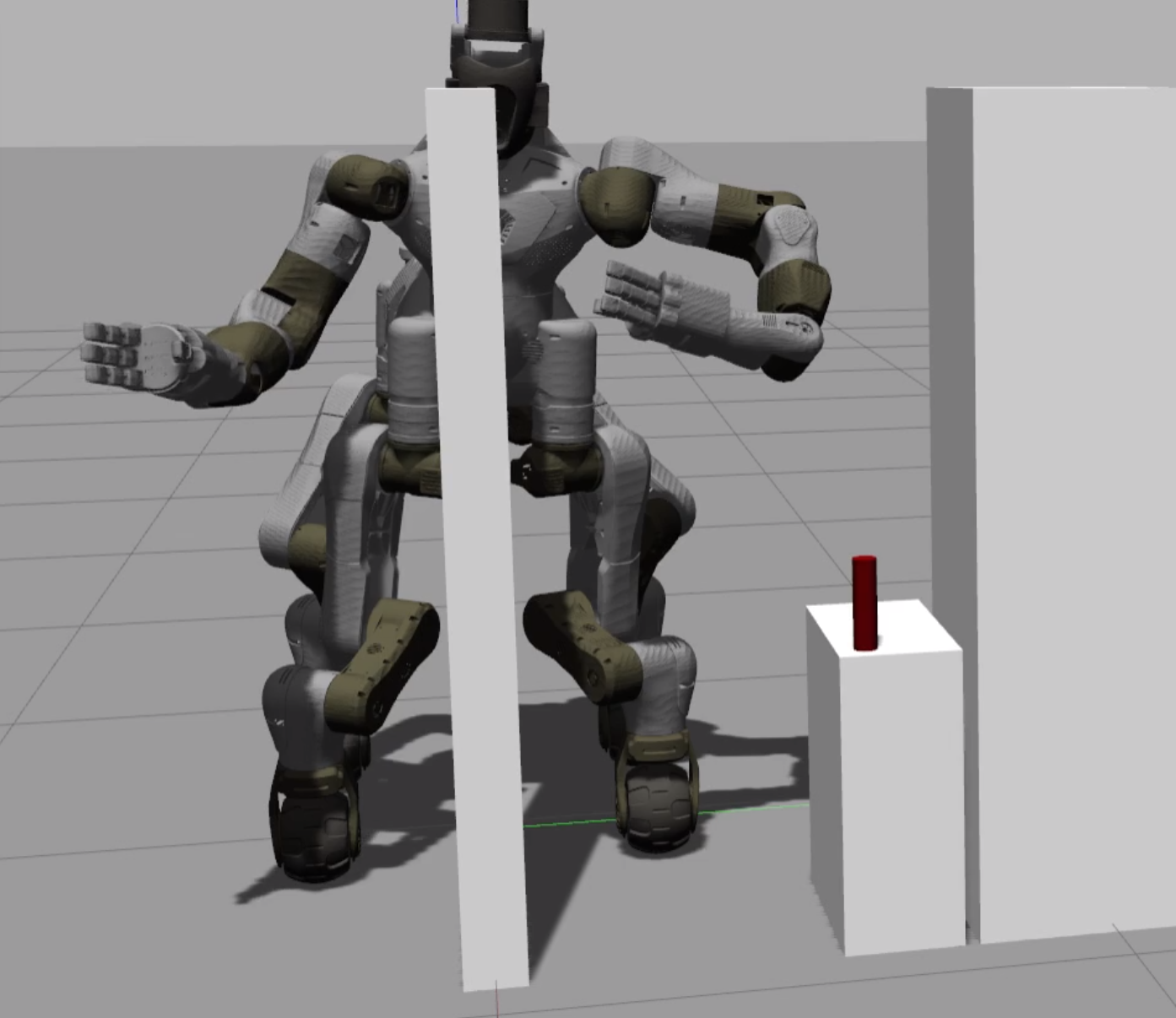}
        \caption{$t_2$}
        \label{fig:subim4}
    \end{subfigure}
    \begin{subfigure}{0.13\textwidth}
        \includegraphics[width=\textwidth]{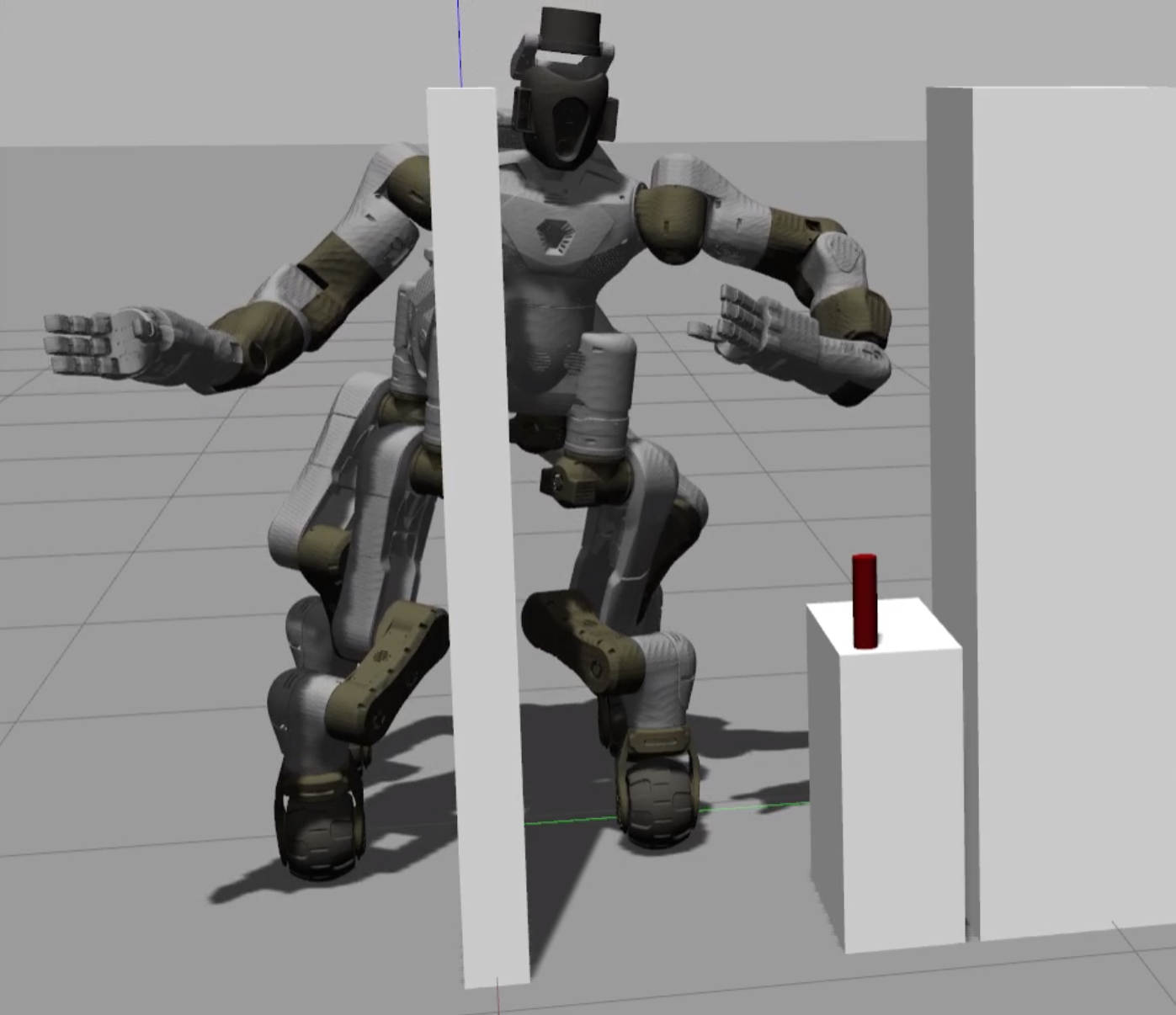}
        \caption{$t_3$}
        \label{fig:subim5}
    \end{subfigure}
    \begin{subfigure}{0.13\textwidth}
        \includegraphics[width=\textwidth]{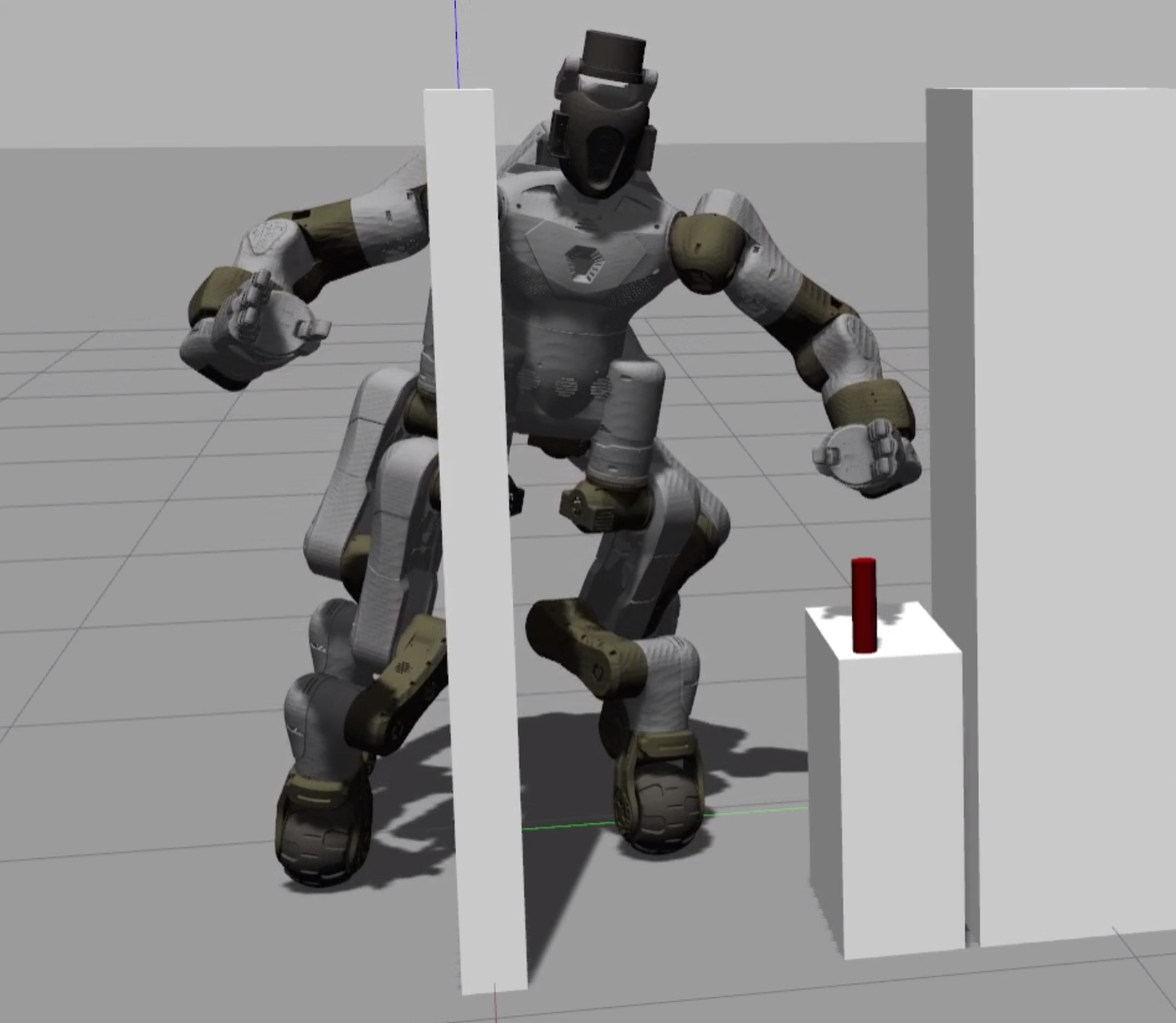}
        \caption{$t_4$}
        \label{fig:subim6}
    \end{subfigure}
    \begin{subfigure}{0.13\textwidth}
        \includegraphics[width=\textwidth]{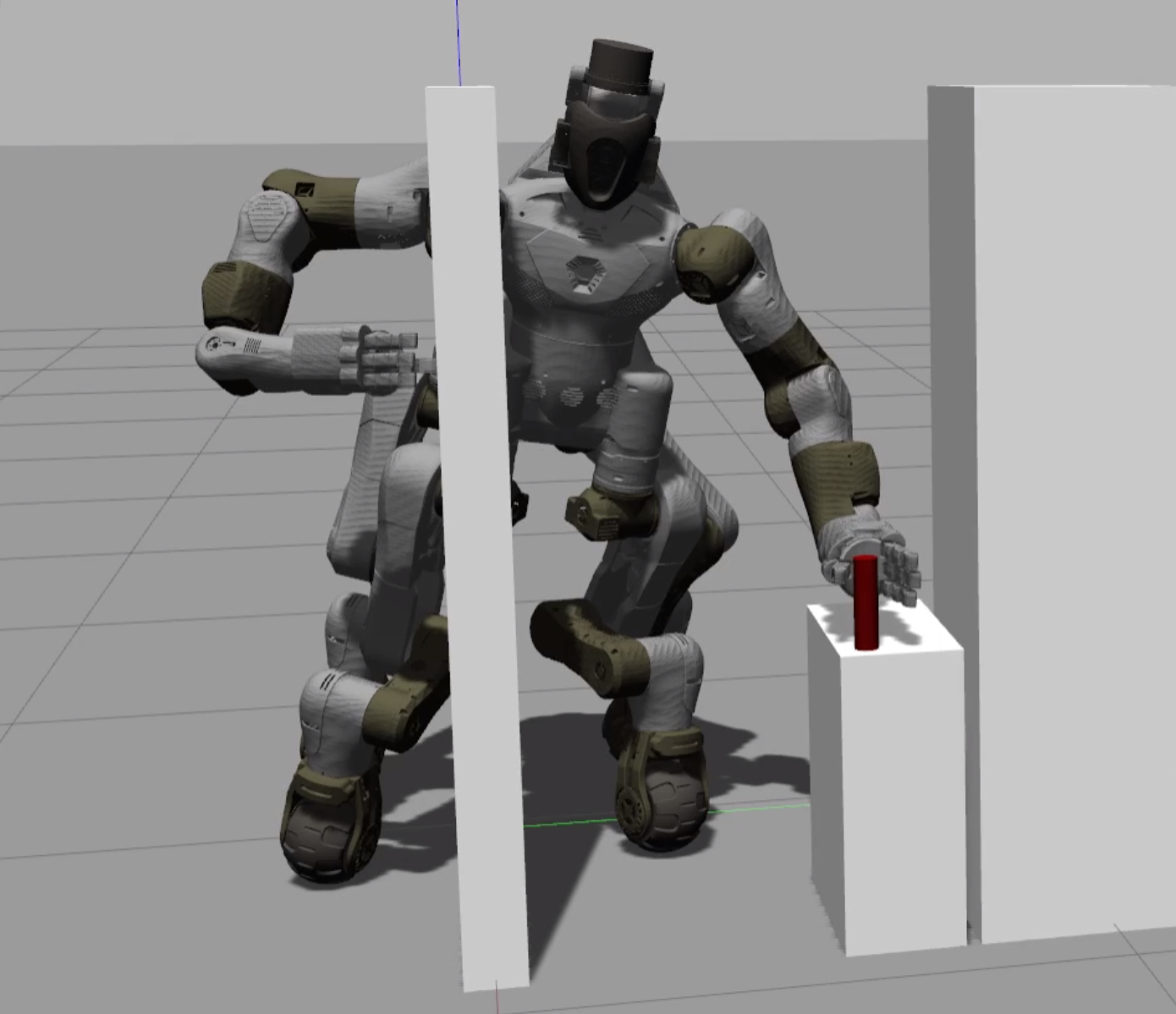}
        \caption{$t_5$ (goal)}
        \label{fig:subim7}
    \end{subfigure}
    \caption{\small Snapshots of whole body motion plan to reach goal pose in the presence of obstacle and satisfying balance constraint.}
    \label{fig:pholus_reach_sim}
\end{figure}

For the balanced-constrained whole-body motion task, we used the Pholus robot. Since MoveIt! utilizes URDF (Unified Robot Description Format) for planning, and it is not possible to change the root link, our planning considered 6-DOF floating-base as root joint, which resulted in 41-dimensional configuration space for whole-body motion planning. To satisfy static balance, we imposed position constraints on the feet to ensure that they stay in contact with the ground and that the center of gravity of the whole body is always inside the support polygon. We collected 20000 stable leg configurations using Bio-IK to train the generative models. The left hand is chosen to reach a bottle in a scenario shown in Figure \ref{fig:pholus_reach_sim} implemented in Gazebo~\cite{Koenig04} physics-based simulation environment, to evaluate the effect of motion planning on static stability. The start pose and the goal pose of the left hand can be seen in Figure \ref{fig:subim1} and \ref{fig:subim7} respectively. It is worth noting that running the same motion planner without sampling from our generative model of the balance constraint manifold (i.e. randomly sampling from free configuration space) results in the robot losing balance and falling to the ground most of the time. 
%Out of XXX runs with limited planning time of XXX sec, the success rates of planning the whole body reach, with and without sampling from our generative model, are XXX\% and YYY\% respectively, showing the effectiveness of our model in generating constraint-satisfying samples without slowing down the planning.

\subsection{ Experiment on Real Robot}
\begin{figure}[h]
    \centering
    \begin{subfigure}{0.15\textwidth}
        \includegraphics[width=\textwidth]{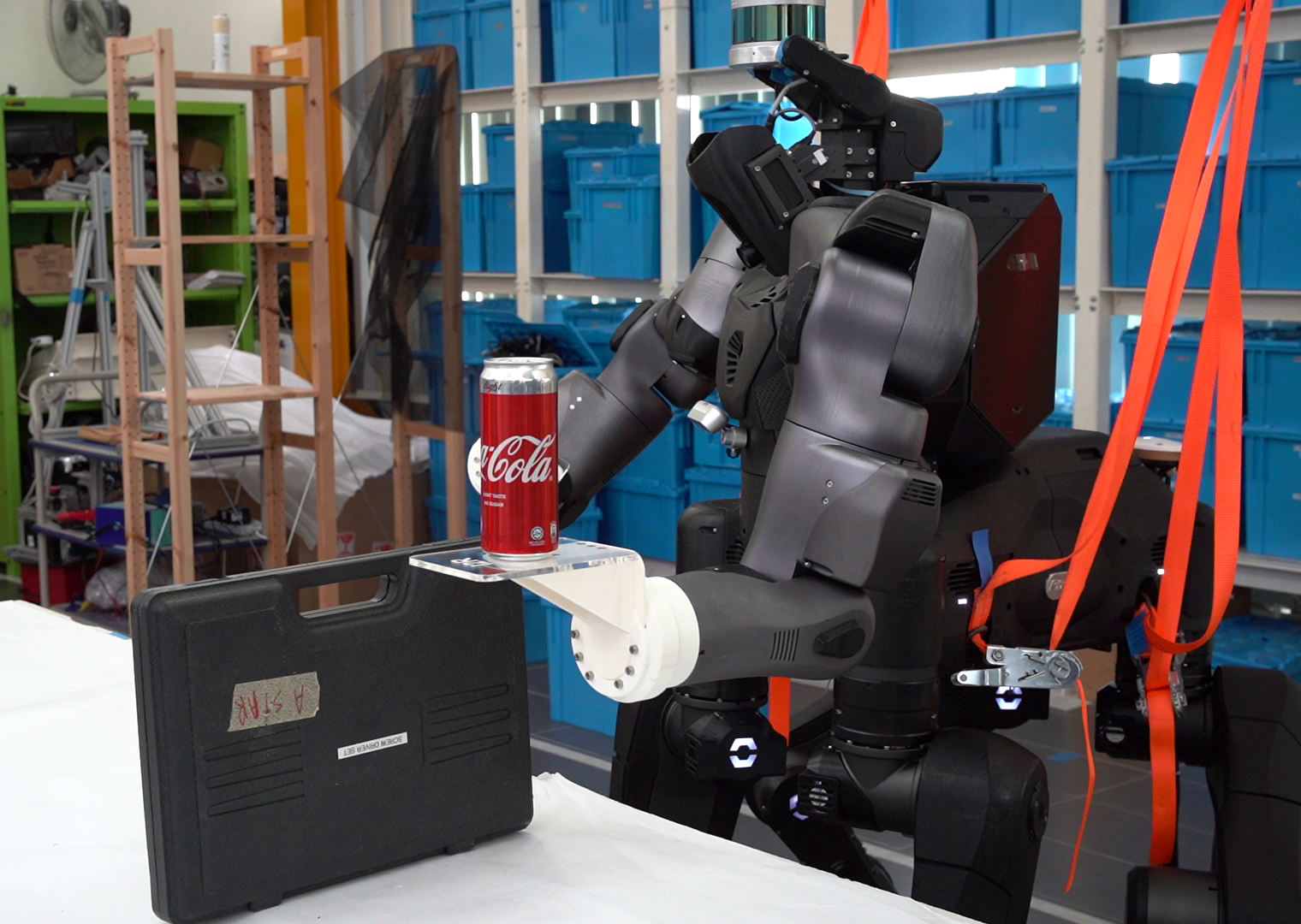} 
        \caption{$t_0$ (initial)}
        \label{fig:real0}
    \end{subfigure}
    \begin{subfigure}{0.15\textwidth}
        \includegraphics[width=\textwidth]{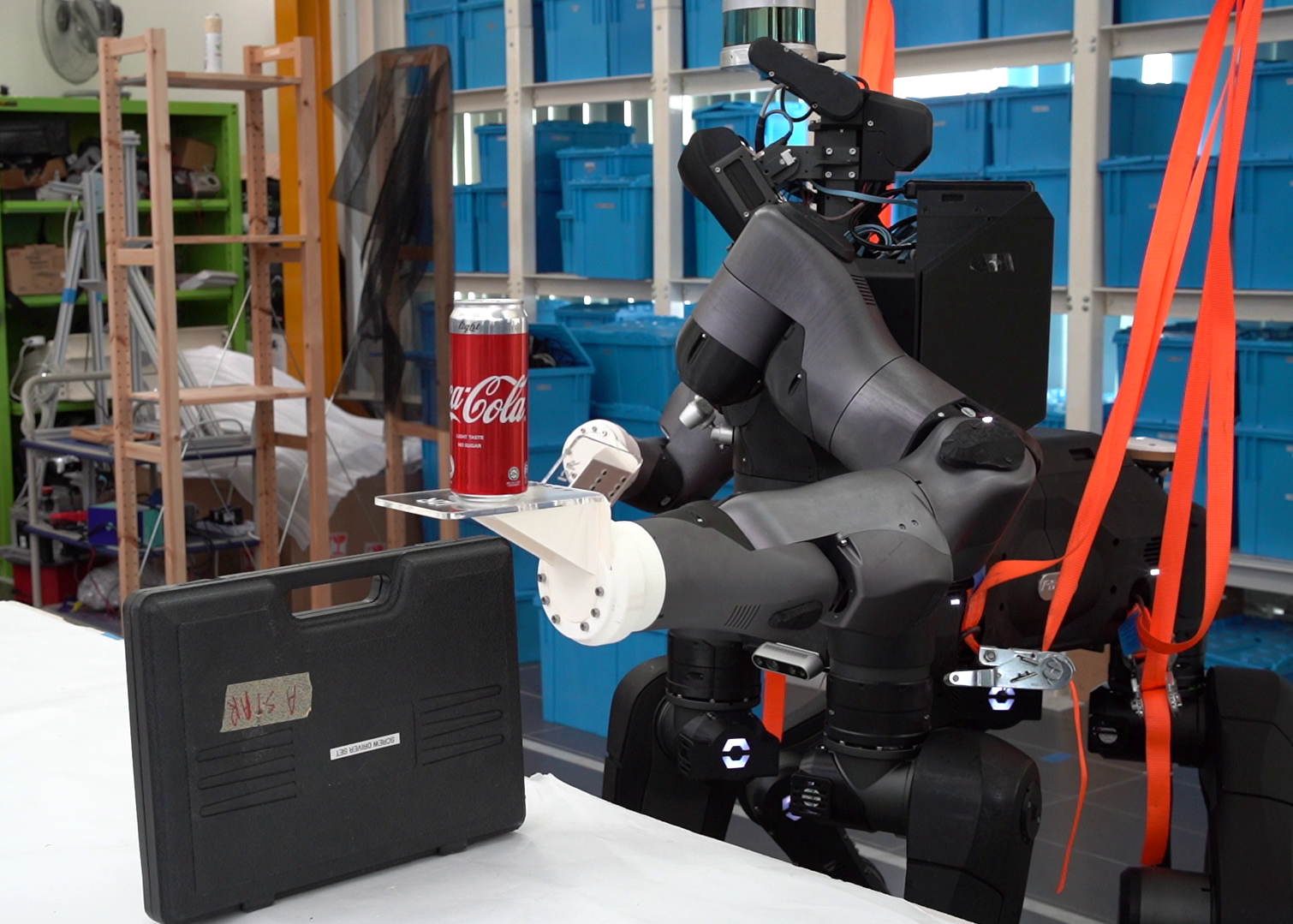} 
        \caption{$t_1$}
        \label{fig:real1}
    \end{subfigure}
    \begin{subfigure}{0.15\textwidth}
        \includegraphics[width=\textwidth]{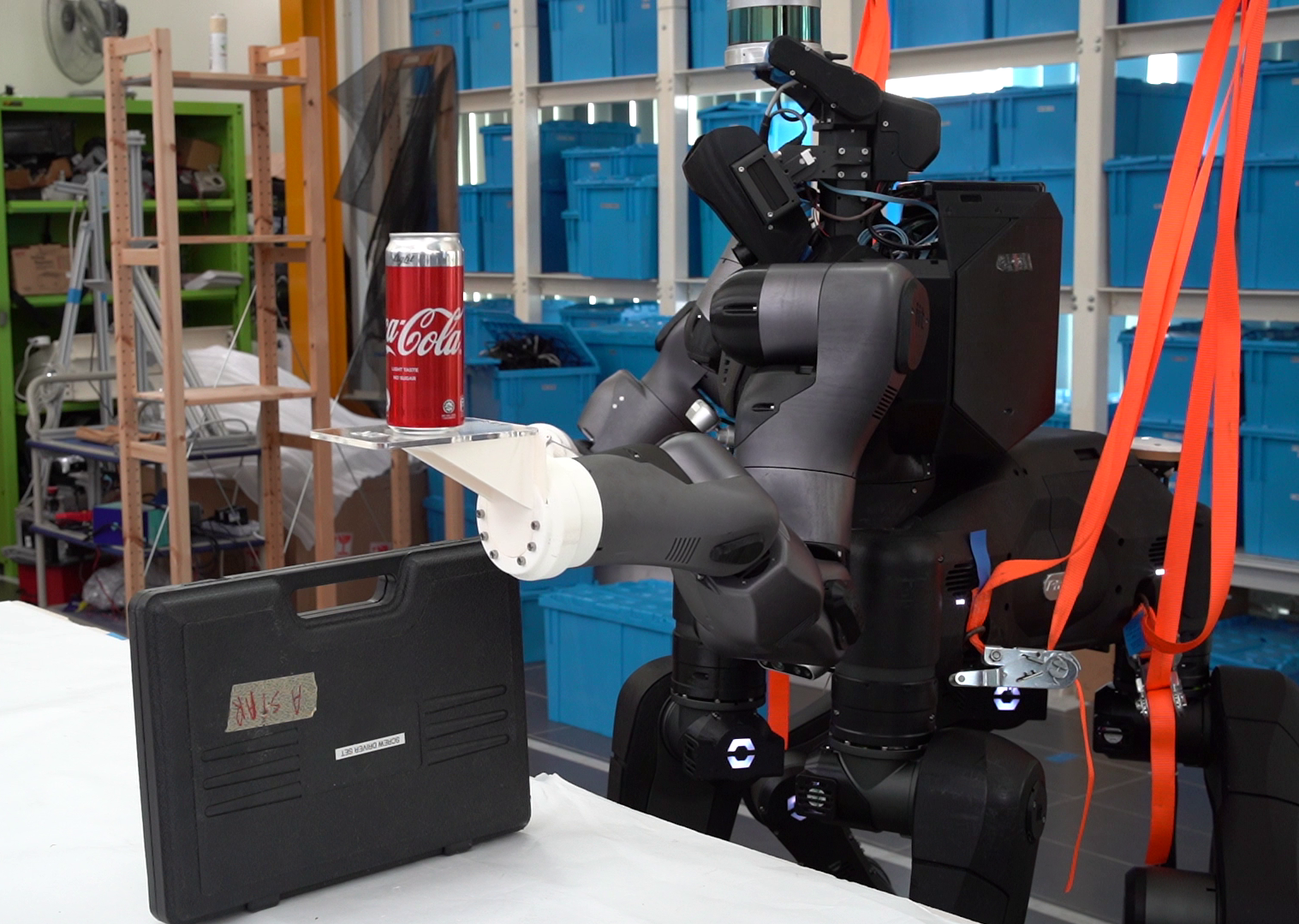}  
        \caption{$t_2$}
        \label{fig:real2}
    \end{subfigure}
    \begin{subfigure}{0.15\textwidth}
        \includegraphics[width=\textwidth]{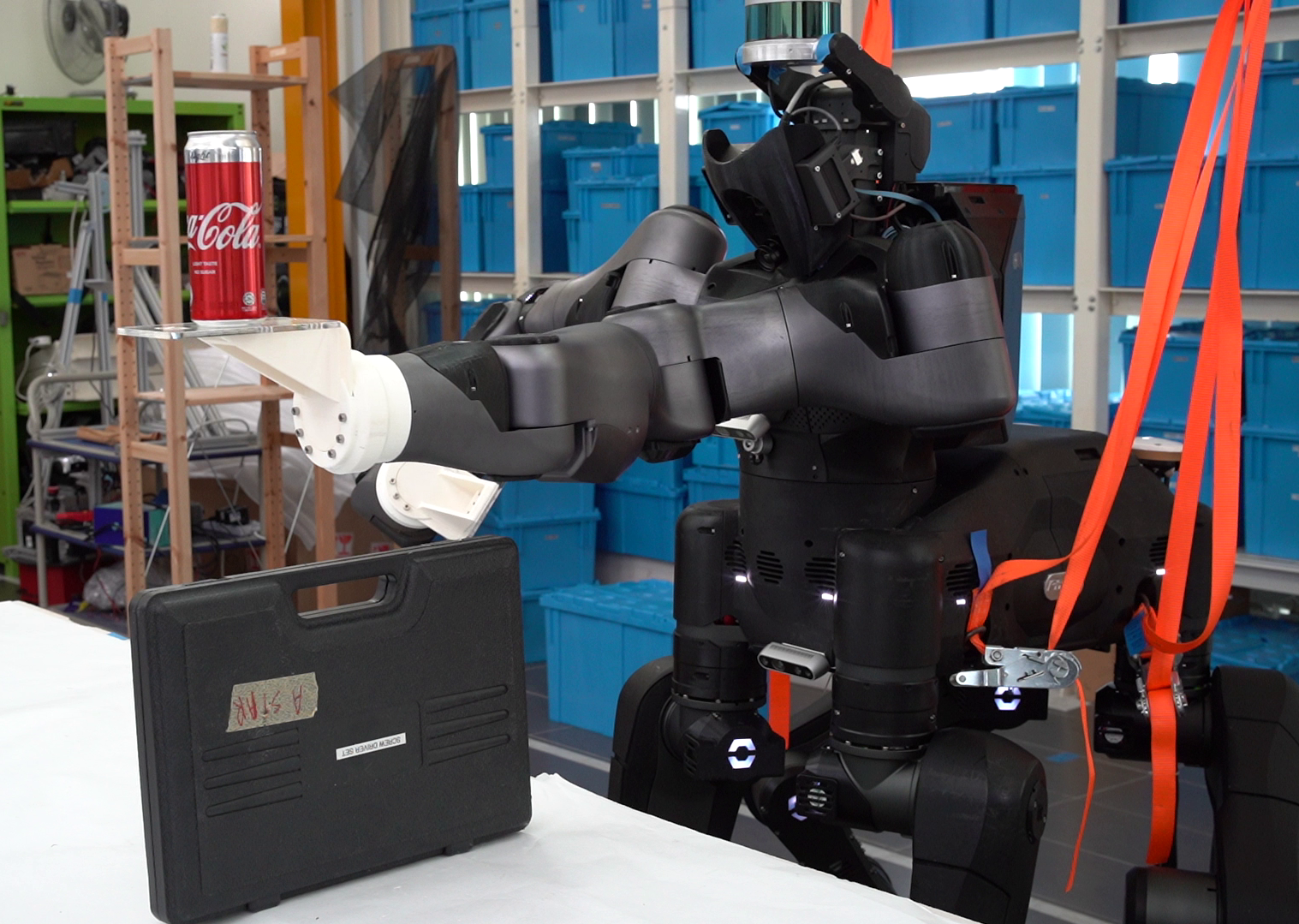}  
        \caption{$t_3$}
        \label{fig:real3}
    \end{subfigure}
    \begin{subfigure}{0.15\textwidth}
        \includegraphics[width=\textwidth]{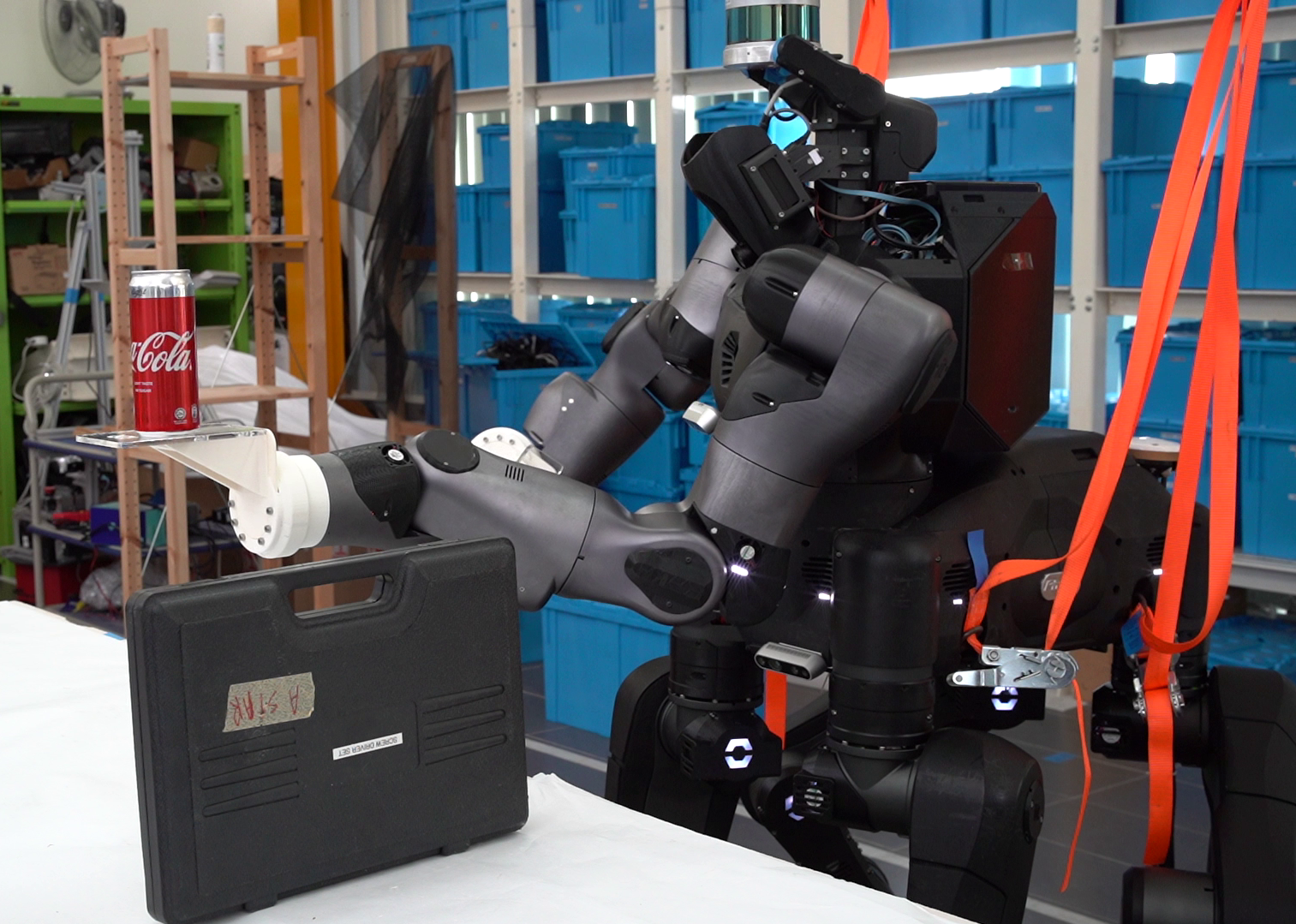}  
        \caption{$t_4$}
        \label{fig:real4}
    \end{subfigure}
    \begin{subfigure}{0.15\textwidth}
        \includegraphics[width=\textwidth]{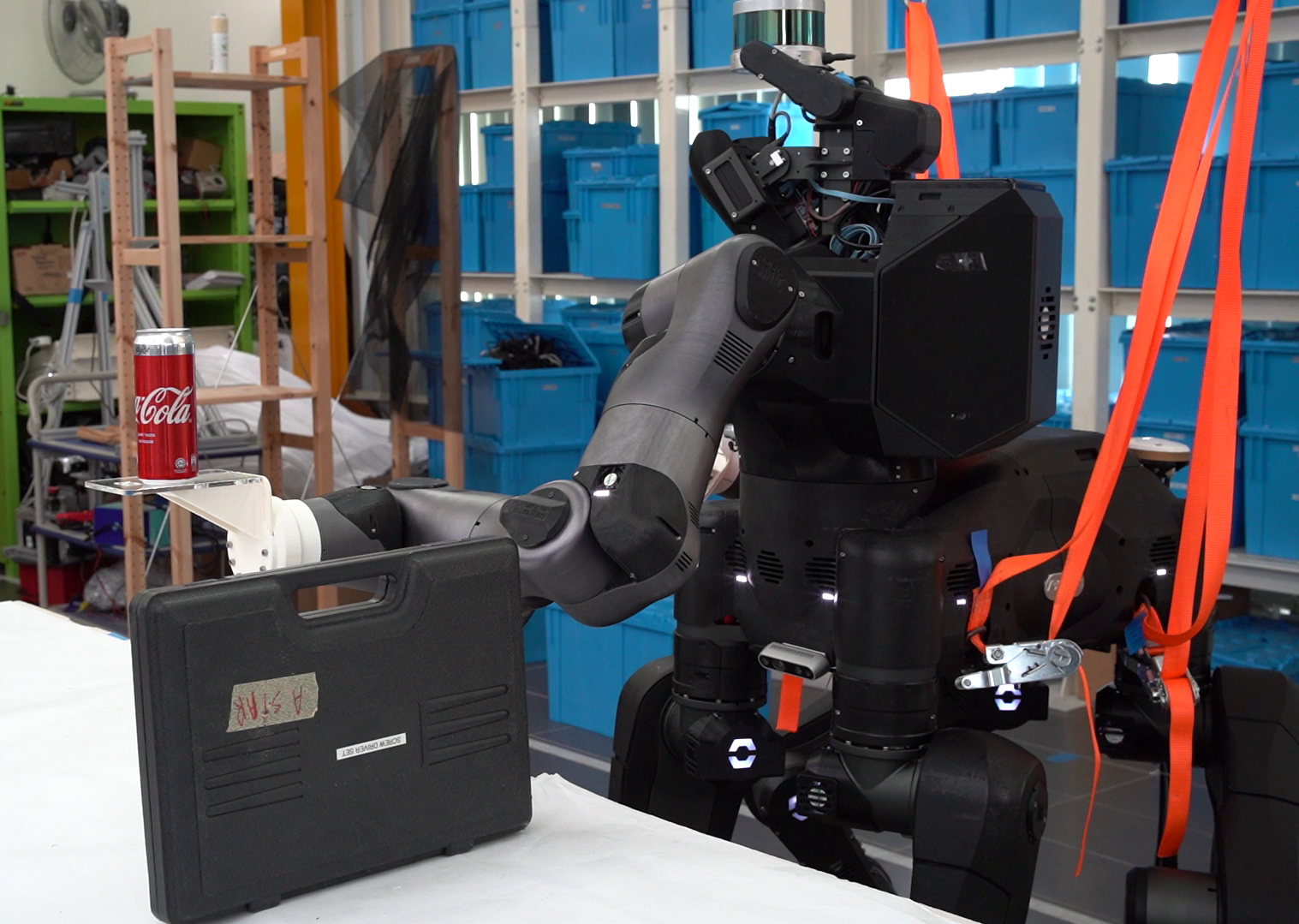}  
        \caption{$t_5$ (goal)}
        \label{fig:real5}
    \end{subfigure}
    \caption{\small Snapshots of real Pholus robot moving an object to target location while maintaining end-effector orientation constraint.}
    \label{fig:real}
\end{figure}

To validate the practicality of the proposed framework, we also implemented our approach on the real Pholus robot. The task involves moving a drink can from an initial location to a target location without spilling the drink (i.e. end-effector orientation constraint in roll, pitch, yaw axes) in the presence of an obstacle which the arm needs to avoid.  For this task, the left arm (7-DOF) and torso (1-DOF) joints were utilized. CVAE and CGAN were conditioned on the parameter representing different yaw angles of the wrist. We collected a total of 10000 configurations for 4 different yaw angles of the wrist using Bio-IK in the simulation environment. Figure \ref{fig:real} shows snapshots of the robot successfully executing a constrained motion plan that moves the drink can over the obstacle and reaches the goal while always maintaining the upright orientation of the can.

 %Even though it is possible to condition the generative model to different yaw angles of the wrist, we consider a fix orientation as seen in Figure \ref{fig:real} in this case. We collected around 5000 different configurations for this orientation using using Bio-IK in simulation environment.    

\section{CONCLUSIONS}
\label{sec:Conclusion}
The main contribution of this work is to introduce generative models for sampling-based constrained motion planning problems, by using them to approximate constraint manifolds embedded in high-dimensional spaces, and generate constraint-satisfying configurations. Our results show that generative models based on CGAN and CVAE are efficient in solving the problem, even if there is significant reduction in the dimensionality of the constraint manifold due to obstacles. They outperform existing precomputed dataset methods, which are unable to generate novel configurations. Furthermore, they can be flexibly used with any sampling-based planner. Even though CGAN gives a higher rate of generating constraint-satisfying samples, we observe that it suffers from the mode collapse issue. CVAE, on the other hand, produces less constraint-satisfying samples (consistent with results in image domain where blurry images are generated) but is better for covering the sampling distribution uniformly. As a result, CVAE gives better overall performance for highly complex and cluttered environments. Another observed advantage of CVAE is that it is easier and more stable to train compared to CGAN.

% Therefore, we conclude that CVAE may be more suitable for the challenging constrained motion planning problems compared to the original CGAN implementation.

% For the future, we plan to include real robot experiments of whole-body constrained motion planning tasks and investigate the  use of generative models on  different constraint tasks.

%Furthermore, between CVAE and CGAN, CVAE gives higher success rates and faster computation for    
 %Our results show that using our generative models with common planners give higher success rates and faster computation than state of the art constrained motion planners.


%%%%%%%%%%%%%%%%%%%%%%%%%%%%%%%%%%%%%%%%%%%%%%%%%%%%%%%%%%%%%%%%%%%%%%%%%%%%%%%%

%%%%%%%%%%%%%%%%%%%%%%%%%%%%%%%%%%%%%%%%%%%%%%%%%%%%%%%%%%%%%%%%%%%%%%%%%%%%%%%%



\section*{ACKNOWLEDGMENT}
{\small We are grateful to Fon Lin Lai, Chong Boon Tan and Samuel Cheong for their help in the simulation and experimental studies.}

% This research is partially supported by the Agency for Science, Technology and Research (A*STAR) under its AME
% Programmatic Funding Scheme (Project \#A18A2b0046).

\bibliographystyle{ieeetr}
\bibliography{root}

\end{document}